%% file: MPG.tex
\documentclass[11pt]{article}   	
\usepackage[margin=1in]{geometry}                		
\usepackage{tipa} 
\usepackage{indentfirst}
\usepackage{graphicx}				
								
\usepackage{amssymb}
\usepackage{color}
\usepackage{subfigure}
\usepackage{float}
\usepackage{amsmath}
\usepackage{booktabs}
\usepackage[flushleft]{threeparttable}
\usepackage{color}
\usepackage{sidecap}
\usepackage{xpatch}
\usepackage[font=small,labelfont=bf]{caption}

\usepackage{colortbl}
\usepackage{mathpazo} 
\usepackage[scaled]{helvet} 
\usepackage{courier} 
\normalfont
\usepackage[T1]{fontenc}
\usepackage{url}
\usepackage[numbers,sort&compress]{natbib}
\usepackage{siunitx} 

\newcommand{\superscript}[1]{\ensuremath{^\textrm{\footnotesize{#1}}}}

\long\def\symbolfootnote[#1]#2{\begingroup%
\def\thefootnote{\fnsymbol{footnote}}\footnote[#1]{#2}\endgroup}
\def\blfootnote{\xdef\@thefnmark{}\@footnotetext}

\usepackage[]{lineno}
\usepackage[colorlinks,
            linkcolor=blue,
            anchorcolor=blue, 
            citecolor=blue, 
            urlcolor=blue,
            ]{hyperref}

\usepackage{graphicx}				
\usepackage{amssymb}
\usepackage{color}
\usepackage{subfigure}
\usepackage{amsmath}
\usepackage{booktabs}
\usepackage{mathrsfs}
\usepackage[flushleft]{threeparttable}
\usepackage{color}
\usepackage{sidecap}
\usepackage{amssymb,bm,mathrsfs}
\usepackage{amsfonts, amsmath}
\usepackage{multirow}
\usepackage{mathpazo} 
\usepackage[scaled]{helvet} 
\usepackage{courier} 
\usepackage{textcomp}
\normalfont
\usepackage[T1]{fontenc}
\usepackage{paralist}

\usepackage{url}

\usepackage[table]{xcolor}

\newcommand{\ourmodel}{MolGNet}
\newcommand{\etal}{\textit{et al}.}

\newcommand{\eg}{\textit{e}.\textit{g}.}

\title{\bf{Learn molecular representations from large-scale unlabeled molecules for drug discovery}}

\author{
 \renewcommand{\thefootnote}{\footnotesize{\arabic{footnote}}}
 Pengyong Li\superscript{1,2},
 Jun Wang\superscript{2}\superscript{,}$^{\footnotesize{*}}$,
 Yixuan Qiao\superscript{2},
 Hao Chen\superscript{2},
 Yihuan Yu\superscript{3},\\
Xiaojun Yao\superscript{4},
 Peng Gao\superscript{2},
 Guotong Xie\superscript{2}\superscript{,}$^{\footnotesize{*}}$,
 and Sen Song\superscript{1}\superscript{,}$^{\footnotesize{*}}$
}

\date{}		

\begin{document}
\maketitle
\symbolfootnote[0]{\hspace{-0.1in} \footnotesize{$^1$} Department of Biomedical Engineering, Tsinghua University, Beijing,
China.}
\symbolfootnote[0]{\hspace{-0.1in} \footnotesize{$^2$} Ping An Technology, Beijing, China.}
\symbolfootnote[0]{\hspace{-0.1in} \footnotesize{$^3$} Beijing University of Chemical Technology, Beijing, China.}
\symbolfootnote[0]{\hspace{-0.1in} \footnotesize{$^4$} College of chemistry and chemical engineering, Lanzhou University, Lanzhou, 730000, China}

\symbolfootnote[0]{\hspace{-0.1in} \footnotesize{$^*$}Corresponding Author. Email: \protect\url{songsen@mail.tsinghua.edu.cn}; \protect\url{junwang.deeplearning@gmail.com};\\
\protect\url{xieguotong@pingan.com.cn }.}

\begin{abstract}
How to produce expressive molecular representations is a fundamental challenge in AI-driven drug discovery. Graph neural network (GNN) has emerged as a powerful technique for modeling molecular data. However, previous supervised approaches usually suffer from the scarcity of labeled data and have poor generalization capability. 
Here, we proposed a novel Molecular Pre-training Graph-based deep learning framework, named MPG, that leans molecular representations from large-scale unlabeled molecules. In MPG, we proposed a powerful \ourmodel~model and an effective self-supervised strategy for pre-training the model at both the node and graph-level. After pre-training on 11 million unlabeled molecules, we revealed that \ourmodel~can capture valuable chemistry insights to produce interpretable representation. The pre-trained \ourmodel~can be fine-tuned with just one additional output layer to create state-of-the-art models for a wide range of drug discovery tasks, including molecular properties prediction, drug-drug interaction, and drug-target interaction, involving 13 benchmark datasets. Our work demonstrates that MPG is promising to become a novel approach in the drug discovery pipeline.

\end{abstract}

\smallskip
\noindent \textbf{Keywords:} molecular representation; deep learning; graph neural network; self-supervised learning

\newpage

\makeatletter
\xpatchcmd{\paragraph}{3.25ex \@plus1ex \@minus.2ex}{3pt plus 1pt minus 1pt}{\typeout{success!}}{\typeout{failure!}}
\makeatother

\input{introduction.tex}
\input{result.tex}

\input{discuss.tex}
\input{method.tex}
\clearpage

\small
\bibliographystyle{naturemag}
\bibliography{ref}

\end{document}

%% file: introduction.tex
\section{Introduction}




Drug discovery is a complicated systematic project spanned over 10-15 years~\cite{hill2012drug}, which is a long journey for a drug from invention to market in practice. Meanwhile, due to the complexity of biological systems and large number of experiments, drug discovery is prone to failure and inherently expensive~\cite{chan2019advancing}. To address these issues, many researchers proposed various computer-aided drug discovery (CADD) methods~\cite{sliwoski2014computational} for small molecule drug design in different stages of early pre-clinical research from hit identification and selection, hit-to-lead optimization, to clinical candidates~\cite{kapetanovic2008computer}. 
Despite the success in assisting drug discovery, traditional CADD methods are mostly based on molecular simulation techniques,  suffering from the huge computation cost and time-consuming procedures, which limits its application in pharmaceutical industry.

The interdisciplinary studies between artificial intelligence (AI)  and drug discovery have received increasing attention due to superior speed and performance. Many AI technologies have been successfully applied in a variety of tasks for drug discovery, such as molecular properties prediction~\cite{ghasemi2018neural}, drug-drug interaction~\cite{ryu2018deep}, and drug-target interaction prediction~\cite{abbasi2020deep, d2020machine}.  One of the fundamental challenges for these studies is how to learn expressive representation from molecular structure~\cite{yang2019analyzing}.
In the early years, molecular representations are based on hand-crafted features such as molecular descriptors or fingerprints~\cite{xue2000molecular}. Most traditional machine learning methods have revolved around feature engineering for these molecular representations. In contrast, there has been a surge of interest in molecular representation learned by deep neural networks, from fitting raw inputs to the specific task-related targets.
Recently, among the promising deep learning architectures, graph neural network (GNN) has gradually emerged as a powerful candidate for modeling molecular data~\cite{gilmer2017neural}.  Because a molecule is naturally a graph that consists of atoms (nodes) connected through chemical bonds (edges), it is ideally suited for GNN. Up to now, various GNN architectures has been proposed~\cite{kipf2016semi,velivckovic2017graph,hamilton2017inductive,gilmer2017neural} and achieved great progress in drug discovery~\cite{Wu2018}. However, there are some limits that need to be addressed. Challenges for deep learning in molecular representation mainly arise from the scarcity of labeled data, as lab experiments are expensive and time-consuming. Thus, training datasets in drug discovery are usually limited in size, and GNNs tends to overfit them, resulting that the learned representations lack of generalizability~\cite{hu2019strategies,rong2020grover}.

One way to alleviate the need for large labeled datasets is to pre-train a model on unlabeled data via self-supervised learning, and then transfer the learned model to downstream tasks~\cite{liu2020self}. These methods have been widely applied and have made a massive breakthrough in computer vision (CV) and natural language processing (NLP)~\cite{krizhevsky2012imagenet,he2020momentum,devlin2019bert}, such as BERT~\cite{devlin2019bert}. Some recent works have employed self-supervised learning to pre-train a model on SMILES~\cite{weininger1988smiles} for learning molecular representation, such as pre-training BERT regarding SMILES as sequences~\cite{honda2019smiles,pesciullesi2020transfer,wang2019smiles,chithrananda2020chemberta}, and pre-training an autoencoder on reconstructing SMILES~\cite{winter2019learning,gomez2018automatic,xu2017seq2seq}. Due to the superior performance of GNN, some researchers began to study the pre-training strategies on molecular graph data~\cite{liu2018ngram,hu2019strategies,rong2020grover}.
However, graph data is often more complicated than image and text data because of the variable topological structures, introducing challenges to adopting a self-supervised learning method to the molecular graph directly. 
Contrastive learning~\cite{liu2020self} is an essential kind of self-supervised approach, which aims at learning to encode what makes two things similar or different. It has achieved great success in learning word representations and visual representations~\cite{chen2020simple,oord2018representation}. Nowadays, some researchers also begin to leverage contrastive learning to empower graph neural networks to learn the representations for graph data from unlabeled input data~\cite{velivckovic2018deep,sun2019infograph,qiu2020gcc}.  Although these contrastive methods achieve great success, most of them are very expensive in computational complexity, which limited their application in pre-training on large-scale datasets like millions of molecules. 
Inspired by language model, many other self-supervised methods for graph has been proposed, such as N-gram~\cite{liu2018ngram}, AttrMasking~\cite{hu2019strategies}, ContextPredict~\cite{hu2019strategies} and MotifPredict~\cite{rong2020grover}. However, these methods mainly focus on node-level representation learning and do not explicitly learn a global graph-level representation, resulting in limited gains in graph-level tasks (e.g., molecular classification). Hu \etal ~\cite{hu2019strategies} employed a supervised molecular property prediction task for pre-training GNN at graph-level, which is limited by the need for large labeled datasets. Moreover, they has confirmed that pre-trained GNNs with pure graph-level or node-level strategy gives limited improvements and sometimes lead to negative transfer on many downstream tasks. Thus, it is desirable to develop an efficient graph-level self-supervised strategy.

To address the above issues, we proposed a novel Molecular Pre-trained Graph-based deep learning framework, named MPG. In MPG, we first developed a novel deep learning network that integrates the powerful capacity of GNN and BERT~\cite{devlin2019bert} to learn molecular representation, called \ourmodel. More importantly, we proposed a  computation-friendly  graph-level self-supervised strategies---Pairwise Subgraph Discrimination (PSD), and combined PSD with AttrMasking~\cite{hu2019strategies} to jointly pre-train our model on the node and graph-level. After pre-training \ourmodel~on 11 million of unlabeled molecules, 
 we first investigated what our model in MPG learned. We found that the pre-trained \ourmodel~can capture meaningful patterns of molecules, including molecular scaffold and some quantum properties, to produce interpretable and expressive representation. Moreover, we conducted extensive experiments to evaluate our MPG on a wide range of drug discovery tasks, including molecular properties prediction, drug-target interaction (DTI) and drug-drug interaction (DDI), with 13 widely used datasets. The experimental results show that our MPG achieved the new state-of-the-art performances on 12 out of 13 datasets, demonstrating the great capacity and generalizability of MPG in drug discovery. In summary, our MPG learns meaningful and expressive molecular representation from large scale unlabeled molecules, and it lays a solid foundation for the application of self-supervised learning in drug discovery pipeline.

%% file: result.tex
\section{Results}
\subsection{The MPG framework}
\begin{figure}[ht]
\centering
\includegraphics[width=1\linewidth]
{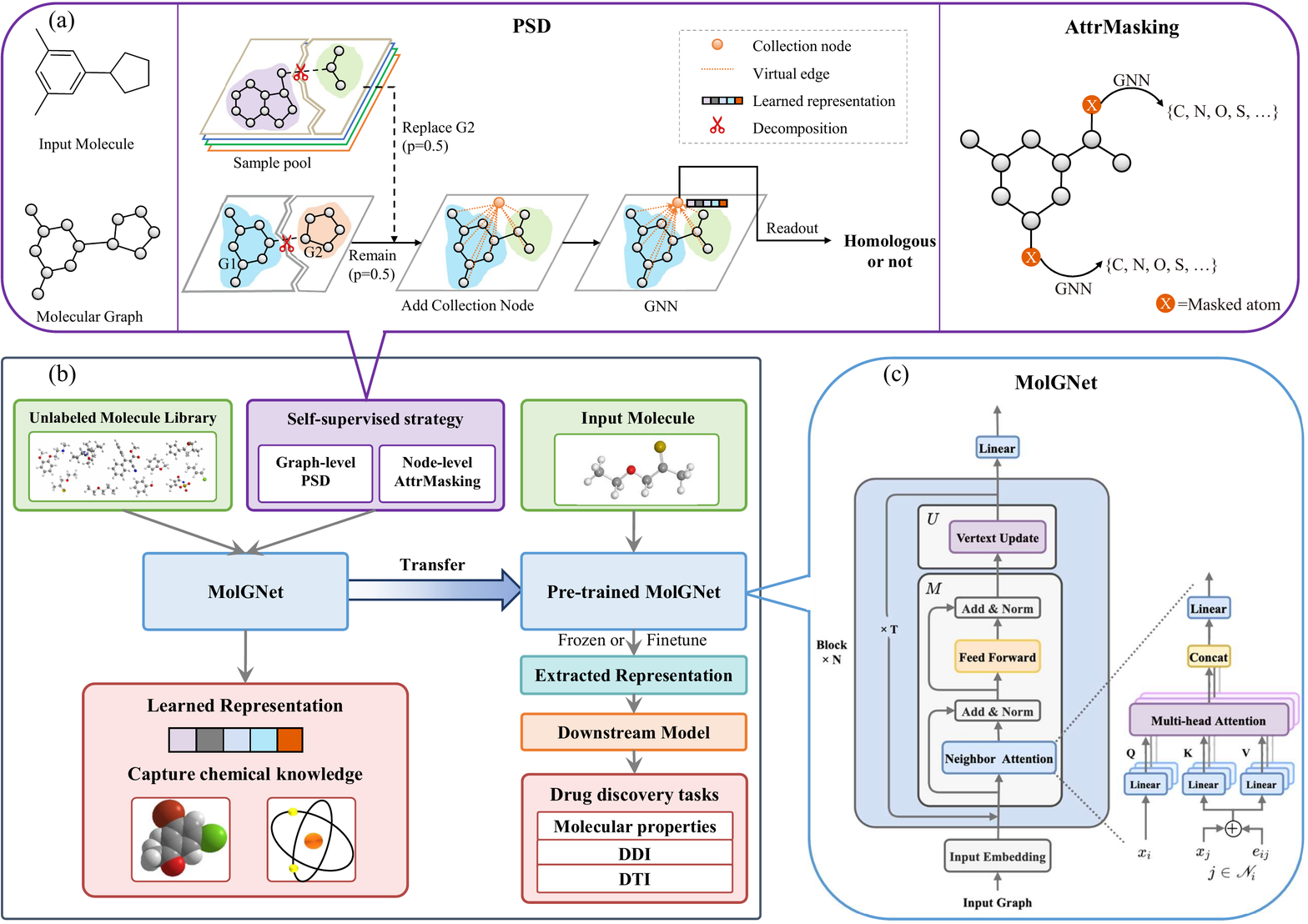}
\caption{\textbf{The overview of MPG framework.} The bottom left sub-figure (b) illustrates the workflow of MPG framework. The MPG framework includes two key components---\ourmodel~and self-supervised strategies. The architecture of \ourmodel~is shown in the bottom right sub-figure (c). The top sub-figure (a) illustrates the schemes of our self-supervised strategy, including PSD we proposed and AttrMasking for pre-training GNN model.}
\label{fig:overview}
\end{figure}

There are two critical aspects to achieving the proposed MPG framework: one is to design a powerful model capable of capturing valuable information from molecular structures; another is to propose an effective self-supervised strategy for pre-training the model. We will introduce the \ourmodel~model and pre-training strategies in MPG (Figure \ref{fig:overview}).

\paragraph{\ourmodel} As shown in Figure \ref{fig:overview} (c),  ~\ourmodel~is composed of a stack of $N=5$ identical layers; each layer performs a shared message passing operation for $T=3$ times recurrently to enable larger receptive fields with less parameters. The message passing operation~\cite{gilmer2017neural} at each time step $t$ contains a message calculation function $\bm M$ and a vertex update function $\bm U$. Formally, these two components work sequentially to update the hidden state $x_i^t$ at each node according to the message passing mechanism. That is

\begin{align}
    \label{eq:mpnn}
    m_i^{t} &= \bm M(\{x_i^{t-1},x_j^{t-1},e_{ij}\},j\in \mathcal{N}_i),\\
    x_i^{t} &= \bm U(h_i^{t-1},m_i^{t}),
\end{align}
where $\mathcal{N}_i$ represents the neighbors of node $i$, $e_{ij}$ denotes the edge between the node $i$ and node $j$, vertex update function $\bm U$ is a gate recurrent unit network (GRU)~~\cite{cho2014learning}, $h_i^{t-1}$ is the hidden state of $\bm U$, and $h_i^0$ is the atom representation $x_i^0$.  Specifically, $\bm M$ has two sub-layers. The first sub-layer conducts the neighbor attention mechanism we proposed, and the second sub-layer is a fully connected feed-forward network. We employ a residual connection around each of the two sub-layers to avoid over-smooth issue~\cite{liu2020towards,li2019deepgcns}, followed by layer normalization.  To facilitate these residual connections, all sub-layers in the model produce outputs of dimension $d=768$. More details about the components of~\ourmodel~can be found in Section \ref{sec:method}.

\paragraph{Self-supervised strategies} Most of the tasks in chemistry (\eg ~molecular properties prediction) crucially rely on globally molecular inherent characteristics.  However, to the best of our knowledge, the current pre-training strategies on molecule graph mainly focus on node-level representation learning~\cite{hu2019strategies,rong2020grover}. Here, we proposed a self-supervised pre-training strategy, named Pairwise Subgraph Discrimination (PSD), that explicitly pre-trains a graph neural network at the graph-level. Inspired by contrastive learning~\cite{liu2020self}, the key idea of PSD strategy (Figure \ref{fig:overview} (a)) is to learn to compare two subgraphs (each decomposed from a graph sample) and discriminate whether they come from the same source (binary classification). In particular, we employ a virtual node, called the collection node, to integrate the information of two subgraphs based on the message passing of GNN. The representation of the collection node, serving as the global representation of the given two subgraphs, learns to predict whether two subgraphs are homologous via maximum likelihood estimation. In order to perform well on the PSD task, it requires the learned collection node representations to encode global information while capable of discriminating the similarity and dissimilarity between pairs of subgraphs. More details about the implementation of PSD can be found in Section \ref{sec:method}. Moreover, we incorporated our PSD strategy with a recently proposed node-level strategy---AttrMasking~\cite{hu2019strategies} for joint pre-training to take full advantage of structural graph information and avoid the negative transfer~\cite{hu2019strategies}. Specifically, AttrMasking is designed to predict the masked node's type, as shown in Figure \ref{fig:overview} (a).

\subsection{MPG captures meaningful patterns of molecules}
\begin{figure}[htbp]
\centering
\includegraphics[width=0.8\linewidth]
{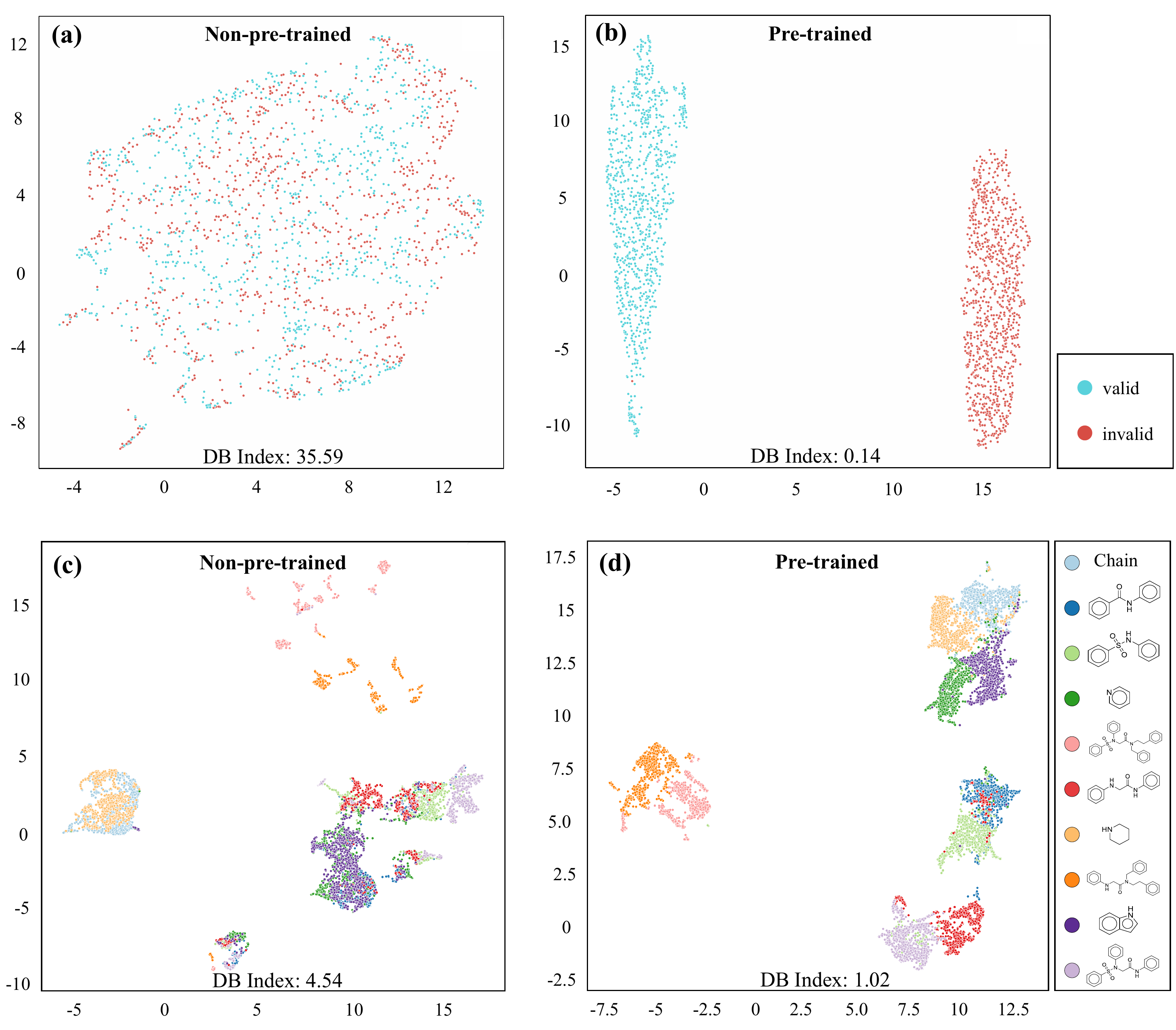}
\caption{\textbf{Visualization of the molecular representation by UMAP.} The molecular representation is the collection node's embedding extracted from the last layer of non-pre-trained or pre-trained \ourmodel. In (a) and (b), pre-trained \ourmodel is capable of distinguishing valid and invalid molecules. In (c) and (d), The different colors represent different scaffolds the molecules belong to. DB index is Davies Bouldin index \cite{dbindex1979}, a lower DB index means that the clustering has a more appropriate separation.}
\label{fig:umap}
\end{figure}


\begin{figure}[ht]
\centering
\includegraphics[width=0.8\linewidth]
{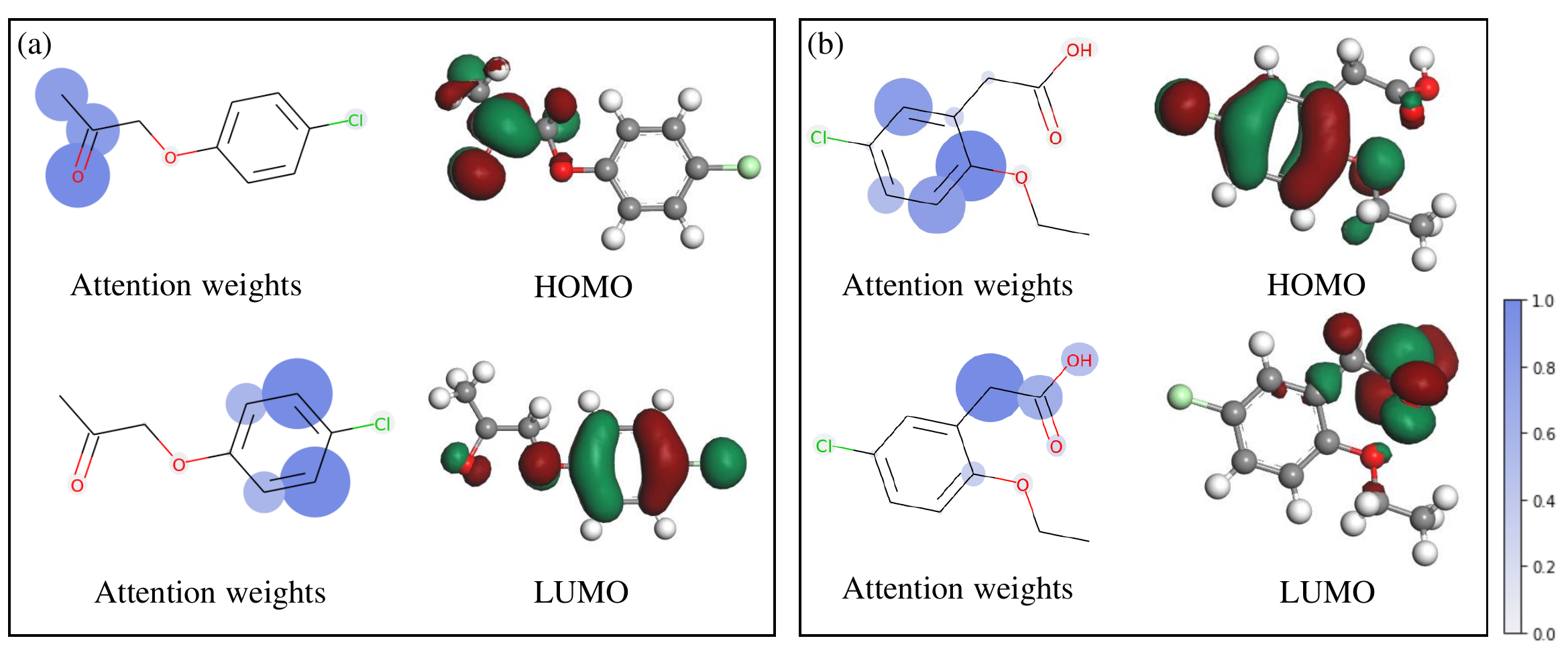}
\caption{\textbf{Two example molecules with attention weights coincided with HOMO and LUMO.} The attention weights represent the importance of atoms to a molecule's global characteristics, extracted from the last layer of \ourmodel~and normalized.  The larger shading size and deeper color both denote larger attention weight. HOMO/LUMO orbitals are calculated by Materials Studio DMol3.} 
\label{fig:homo}
\end{figure}

To pre-train the \ourmodel~in MPG, we first constructed a large-scale dataset that contains 11 million molecules from ZINC\cite{sterling2015} and ChemBL\cite{gaulton2011chembl} dataset. To preserve the diversity, the filtered molecules from ZINC cover a wide range of molecular weight (>=200 Daltons) and LogP (>=-1). Each molecule is represented by a set of atom features and a set of bond features (Table S1). We leveraged AttrMasking and our PSD strategies to train the \ourmodel~jointly. The hyperparameter settings and the learning curves are listed in Figure S1 and Table S2, respectively. After pre-training, we attempted to test whether the pre-trained \ourmodel~can learn the intrinsic patterns underlying the global molecular characteristic or the interactions between atoms.


 To intuitively observe what features the model learned, we visualized the representation extracted by the pre-trained models and tried to explore whether the molecular representations derived from our model hint at chemistry knowledge. First, we investigated  whether MPG can discriminate the valid molecules from the invalid molecules by their structures, which is the most basic ability for a chemist. The invalid molecular structures conflict with the standard chemical knowledge, such as incorrect valence for atoms. Here, we randomly select 1000 molecules from the ZINC dataset and disturb the molecular structures to produce the invalid molecules by shuffling atom features. For each valid and invalid molecule, we extracted the collection node's embedding from the last layer of pre-trained \ourmodel as the molecular representation. Once obtained, the representations of both valid and invalid molecules are visualized in the projected 2D space by uniform manifold approximation and projection (UMAP)~\cite{mcinnes2018umap}. We also performed the same analysis on the \ourmodel~model that was not pre-trained for comparison. As shown in Figure \ref{fig:representation} (a) and (b), non-pre-trained \ourmodel~shows no obvious cluster, and the molecules overlap in a mess without meaningful patterns. After pre-training, the model separated the molecules with two distinct clusters corresponding to valid and invalid molecules (The DB index \cite{dbindex1979} was decreased from 33.59 to 0.14, indicating a more appropriate separation), demonstrating that the pre-trained model can identify whether the molecule is valid.
 
Second, we tested whether MPG can encode the scaffold information from molecular structure. The scaffold is an essential concept in chemistry to represent the core structure of a molecule, which provides a basis for systematic investigations of molecular cores and building blocks~\cite{bemis1996properties,hu2016computational}. Here, we visualized the representation of the molecules with different scaffolds by UMAP.  Specifically, we chose ten most common scaffolds from the ZINC dataset and randomly sampled 1000 molecules for each selected scaffold, resulting in 10000 molecules labeled with ten different scaffolds. Similarly, the collection node's embedding is regarded as the representation for the molecule.
Figure \ref{fig:umap} (c) and (d) shows the distributions of the representations of molecules produced by the \ourmodel~with or without the pre-training scheme. Compared with the non-pre-trained \ourmodel, the pre-trained \ourmodel~shows more distinctive clusters corresponding to the ten molecular scaffolds. It indicates that the pre-trained model is capable of capturing globally inherent molecular characteristics. This capacity may be because our PSD strategy prompts \ourmodel~to perceive global structural insights or chemical rules, which identifies the scaffolds to accurately discriminate whether two sub-graphs are homologous. It should be noted that the molecules with different scaffolds usually have very different properties. Thus our MPG could provide high-quality representations for the downstream tasks.

 
Finally, we investigate the interpretation of MPG in a more fine-grained way. We colored each atom of selected molecules with the attention weights on the collection node obtained from the last layer of the pre-trained \ourmodel. The attention weights represent the contribution of atoms to the global feature. To see whether these attention scores are near related to the critical structural factor of molecules or not, we plotted the Highest Occupied Molecular Orbital (HOMO) and the Lowest Unoccupied Molecular Orbital (LUMO) obtained from density functional theory (DFT) calculations for molecules.  Surprisingly, we could always find some heads whose attention weights coincided with the areas to which the HOMO and LUMO are distributed (Figure \ref{fig:homo}). HOMO and LUMO represent the energy required to extract or inject an electron from/to a molecule, respectively, which have crucial effects on the molecular properties, such as redox ability, optical properties, and chemical reactivity. In summary, our \ourmodel~can leverage valuable chemistry knowledge to guide the generation of molecular representation. 

Overall, MPG has been proved to be capable of learning interpretable molecular representations that capture some common sense in chemistry, which might bridge the gap between the pre-training and downstream tasks to boost performances.

\subsection{MPG advances the state-of-the-art in molecular properties prediction}

\definecolor{}{rgb}{0.88,1,1}
\definecolor{LightGreen}{rgb}{0.88,1,0.88}
\definecolor{LightRed}{rgb}{1,0.88,0.88}
\definecolor{LightGray}{gray}{0.8}

\begin{table}[ht]

    \centering
    \caption{The performance comparison on molecular properties prediction.}

 \resizebox{1\textwidth}{!}{
   \begin{threeparttable}
\renewcommand{\arraystretch}{1.1}
\begin{tabular}{ccccccc|ccc}
\toprule
\multicolumn{7}{c|}{Classification (AUC-ROC)}&\multicolumn{3}{c}{Regression (RMSE)} \\
\hline
Dataset & {BBBP} & {SIDER} & {ClinTox} & {BACE} & {Tox21} & {ToxCast}  & {FreeSolv} & {ESOL} & {Lipo} \\
\# Molecules & {2039} & {1427} & {1478} & {1513} & {7831} & {8575}& {642} & {1128} & {4200}  \\
\hline
TF\_Robust \cite{ramsundar2015massively} & $0.860_{(0.087)}$ & $0.607_{(0.033)}$ & $0.765_{(0.085)}$ & $0.824_{(0.022)}$ & {$0.698_{(0.012)}$} & $0.585_{(0.031)}$& $4.122 _{(0.085)}$ & $1.722 _{(0.038)}$ & $0.909 _{(0.060)}$ \\
GraphConv \cite{kipf2016semi}& $0.877_{(0.036)}$ & $0.593_{(0.035)}$ & $0.845_{(0.051)}$ & $0.854_{(0.011)}$ & {$0.772_{(0.041)}$} & $0.650_{(0.025)}$ & $2.900_{(0.135)}$ & $1.068_{(0.050)}$ & $0.712_{(0.049)}$ \\
Weave  \cite{kearnes2016molecular}& $0.837_{(0.065)}$ & $0.543_{(0.034)}$ & $0.823_{(0.023)}$ & $0.791_{(0.008)}$ & {$0.741_{(0.044)}$} & $0.678_{(0.024)}$& $2.398_{(0.250)}$ & $1.158_{(0.055)}$ & $0.813_{(0.042)}$ \\
SchNet \cite{schutt2017schnet}& $0.847_{(0.024)}$ & $0.545_{(0.038)}$ & $0.717_{(0.042)}$ & $0.750_{(0.033)}$ & {$0.767_{(0.025)}$} & $0.679_{(0.021)}$& $3.215_{(0.755)}$ & $1.045_{(0.064)}$ & $0.909_{(0.098)}$  \\
MPNN  \cite{gilmer2017neural}& $0.913_{(0.041)}$ & $0.595_{(0.030)}$ & $0.879_{(0.054)}$ & $0.815_{(0.044)}$ & {$0.808_{(0.024)}$} & $0.691_{(0.013)}$& $2.185_{(0.952)}$ & $1.167_{(0.430)}$ & $0.672_{(0.051)}$  \\
DMPNN \cite{yang2019analyzing}&  $0.919_{(0.030)}$ &  $0.632_{(0.023)}$ & $0.897_{(0.040)}$ & $0.852_{(0.053)}$ & { $0.826_{(0.023)}$} &  $0.718_{(0.011)}$& $2.177_{(0.914)}$ & $0.980_{(0.258)}$ & $0.653_{(0.046)}$ \\
MGCN  \cite{lu2019molecular}& $0.850_{(0.064)}$ & $0.552_{(0.018)}$ & $0.634_{(0.042)}$ & $0.734_{(0.030)}$ & {$0.707_{(0.016)}$} & $0.663_{(0.009)}$& $3.349_{(0.097)}$ & $1.266_{(0.147)}$ & $1.113_{(0.041)}$ \\
AttentiveFP \cite{xiong2019pushing}& $0.908_{(0.050)}$ & $0.605_{(0.060)}$ &  $0.933_{(0.020)}$ &  $0.863_{(0.015)}$ & {$0.807_{(0.020)}$} & $0.579_{(0.001)}$ &  $2.030_{(0.420)}$ &  $0.853_{(0.060)}$ &  $0.650_{(0.030)}$ \\
\hline
\cellcolor{LightGray}N-GRAM \cite{liu2019n}& $0.912_{(0.013)}$ & $0.632_{(0.005)}$ & $0.855_{(0.037)}$ &  $0.876_{(0.035)}$ & {$0.769_{(0.027)}$} & - &$2.512_{(0.190)}$ & $1.100_{(0.160)}$ & $0.876_{(0.033)}$ \\
\cellcolor{LightGray}Smiles Transformer\cite{honda2019smiles} & $0.900_{(0.053)}$ & $0.559_{(0.017)}$ & $0.962_{(0.064)}$ & $0.719_{(0.023)}$ & {$0.706_{(0.021)}$} & - &  ${2.246}_{(0.237)}$ & ${1.144}_{(0.118)}$ &  ${1.169}_{(0.031)}$ \\
\cellcolor{LightGray}HU. et.al.~\cite{hu2019strategies} & $0.915_{(0.040)}$ & $0.614_{(0.006)}$ & $0.762_{(0.058)}$ & $0.851_{(0.027)}$ & {$0.811_{(0.015)}$} & $0.714_{(0.019)}$& - & - & - \\

\cellcolor{LightGray}GROVER~\cite{rong2020grover}& 
$\bm {0.940}_{(0.019)}$ & 
${0.658}_{(0.023)}$ & 
${0.944}_{(0.021)}$ & 
${0.894}_{(0.028)}$ & {${0.831}_{(0.025)}$} & ${0.737}_{(0.010)}$ &  ${1.544}_{(0.397)}$ & ${0.831}_{(0.120)}$ &  $0.560_{(0.035)}$\\
\hline
\cellcolor{LightGray}MPG & 
$0.922{(0.012)}$ & 
$\bm{0.661}_{(0.007)}$ & 
$\bm{0.963}_{(0.028)}$ &
$\bm{0.920}_{(0.013)}$ & 
{$\bm{0.837}_{(0.019)}$} & $\bm{0.748}_{(0.005)}$& 
 $\bm{1.269}_{(0.192)}$ &
 $\bm{0.741}_{(0.017)}$ & 
 $\bm{0.556}_{(0.017)}$ \\

\bottomrule
\end{tabular}%

   \begin{tablenotes}
   \footnotesize
  \item The methods in shading cells are pre-trained methods.
 Each dataset was split into train/validation/test set by the scaffold split with a ratio of 8:1:1. We conduct three runs on three random seeded scaffolds splitting and reported the mean along with standard deviation (the numbers in brackets) on the test set. The baselines' performances are taken from GROVER~\cite{rong2020grover} and SMILES Transformer~\cite{honda2019smiles}.
 \end{tablenotes}
  \end{threeparttable}
}
\vspace{-2ex}

\label{tab:performance_res}
\end{table}

Quantitative structure-activity relationship (QSAR) analysis, aiming at screening large libraries of molecules with desired properties, has emerged as a powerful computational approach in drug discovery~\cite{cherkasov2014qsar}. 
This section comprehensively evaluates our MPG on nine widely used datasets covering various molecular properties, including physical chemistry, biophysics, and physiology properties.  Details about data sets are referred to Supplementary Information (SI). To offer a fair comparison, we followed the same experimental setting as previous best method---GROVER~\cite{rong2020grover}.  We added a randomly-initialized linear classifier on top of the graph-level representations obtained by our pre-trained \ourmodel, and fine-tuned the model using the training sets of downstream task datasets. The optimal hyperparameter settings and learning curves are listed in Table S3 and Figure S2, respectively.

Table \ref{tab:performance_res} summarizes the results that compare MPG with previous self-supervised methods and supervised methods on molecular properties prediction. It indicates that our MPG achieves state-of-the-art performance on 8 out of 9 data sets. Compared to previous best methods--GRVOVER, the overall improvement is 13.9\% (0.9\% on classification tasks and 26.9\% on regression tasks). Meanwhile, GROVER contains 100 million parameters, while \ourmodel ~contains 53 million parameters. Better performance with less parameters demonstrates the effectiveness of our MPG. In particular, MPG achieved larger gains on small data sets, such as ClinTox, BACE, FreeSolv, and ESOL, confirming that our MPG can boost the performance on the tasks with very few labeled data. These superior performances could be attributed to the self-supervised strategy we proposed. The self-supervised strategy in GROVER only focuses on local structure learning. In contrast, our strategy enables the model to capture more valuable information at both node and graph-level.

\begin{figure}[ht]
\centering
\includegraphics[width=0.6\linewidth]
{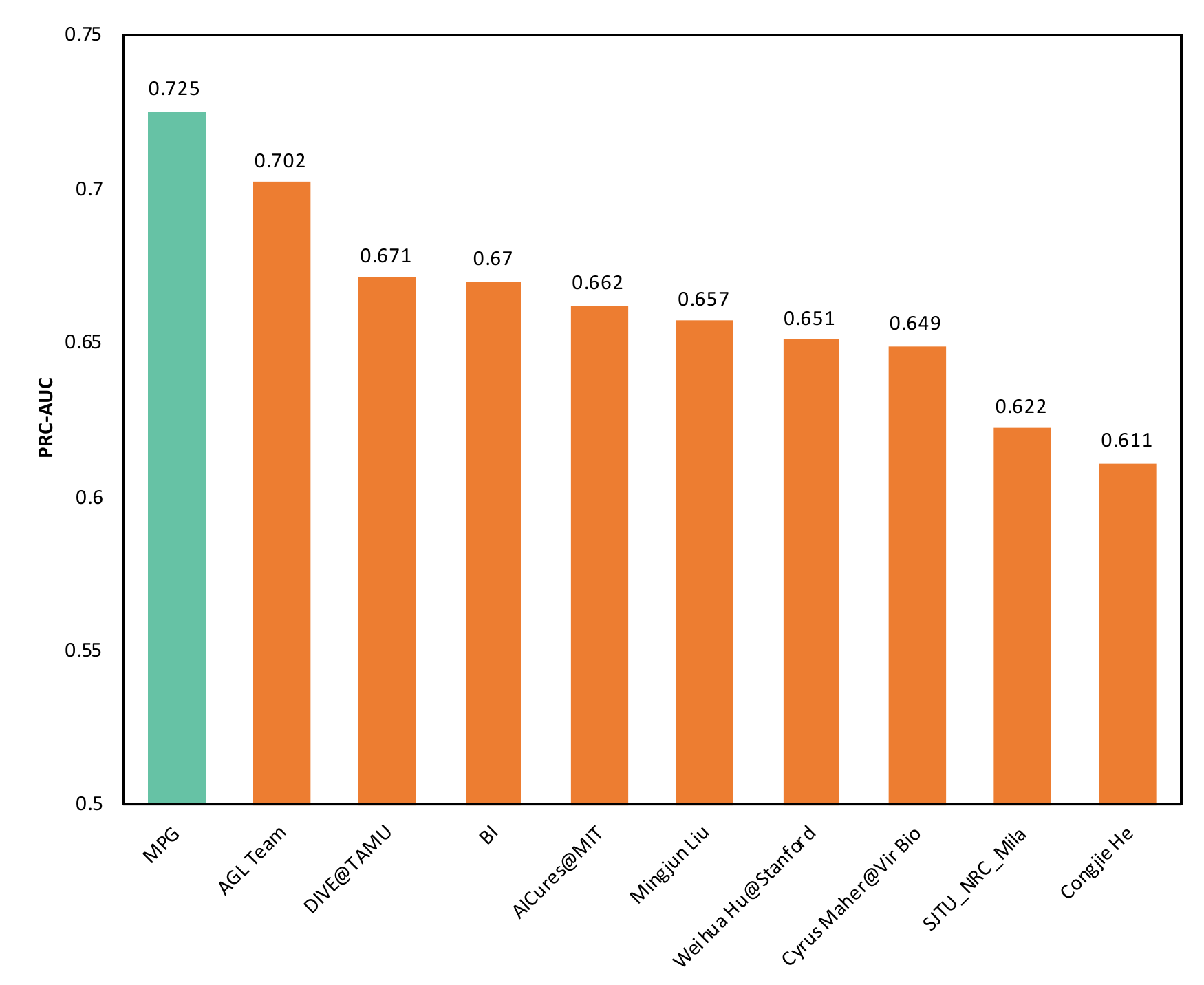}
\caption{MPG outperforms other methods in the antibacterial properties prediction open task hosted by MIT J-Clinic.}
\label{fig:cure}
\end{figure}

Inspired by the impressive performance on molecular properties prediction, we took part in an open task released by MIT J-Clinic recently (\protect\url{https://www.aicures.mit.edu/tasks}), aiming at predicting antibacterial properties of molecules on \textit{Pseudomonas aeruginosa} datasets, for the treatment of secondary infections in patients with  COVID-19.
Our MPG currently ranks the first with PRC-AUC of 0.725 on this benchmark, outperforming the runner-up with an improvement of 3.3\% (Figure \ref{fig:cure}). It is an inspiring real-world application of MPG, making it possible to find promising drugs for fighting COVID-19 and other emerging pathogens. It can decrease the healthcare burden of secondary infections and increase the likelihood of survival of critically ill patients with COVID-19.




\subsection{MPG predicts the drug-drug interaction accurately  and rationally}

\begin{figure}[ht]
\centering
\includegraphics[width=0.5\linewidth]
{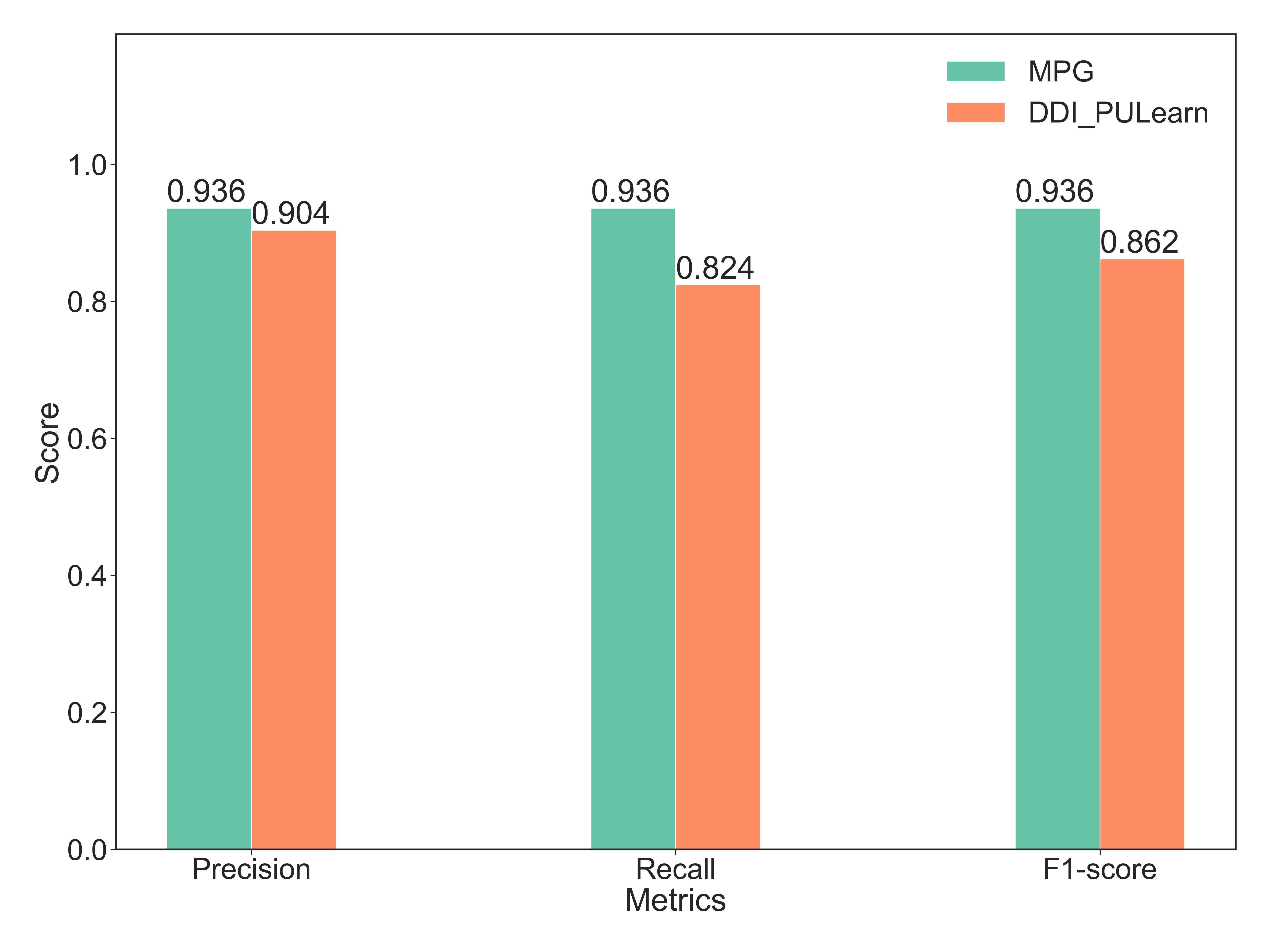}
\caption{5-fold cross-validation classification performance on TWOSIDES dataset.}
\label{fig:twosides}
\end{figure}

\begin{table*}[ht]
\centering
    \caption{MPG provides more accurate DDI prediction than other strong baselines on BIOSNAP dataset.}
    \label{DDI_pred_figure}
      \begin{threeparttable}
    \begin{tabular}{llcccc}
    \toprule
    Model \ \ \ \ \ & ROC-AUC & PR-AUC & F1  \\ \hline 
    \multirow{1}{*}{LR} & $0.802_{(0.001)}$ & $0.779 _{(0.001)}$& $0.741_{(0.002)}$ \\
   \hline
    \multirow{1}{*}{Nat.Prot~\cite{vilar2014similarity}}  & $0.853_{(0.001)}$ & $0.848 _{( 0.001)}$& $0.714_{(0.001)}$\\
    \hline
    \multirow{1}{*}{Mol2Vec~ \cite{jaeger2018mol2vec}}  & $0.879_{(0.006)}$ & $0.861 _{( 0.005)}$& $0.798_{(0.007)}$ \\
    \hline
    \multirow{1}{*}{MolVAE~ \cite{gomez2018automatic}}  &  $0.892_{(0.009)}$ & $0.877 _{( 0.009)}$& $0.788_{(0.033)}$  \\
   \hline
    \multirow{1}{*}{DeepDDI~\cite{ryu2018deep}} & $0.886_{(0.007)}$ & $0.871_{( 0.007)}$& $0.817_{(0.007)}$ \\
   \hline
     \multirow{1}{*}{CASTER~\cite{huang2020caster}}  &  $0.910 _{( 0.005)}$ &  $0.887 _{( 0.008)}$ &  $0.843 _{( 0.005)}$  \\
        \hline
     \multirow{1}{*}{MPG}  & $\bm{0.966} _{( 0.004)}$& $\bm{0.960}_{( 0.004)}$ & $\bm{0.905}_{( 0.008)}$  \\
    
    \bottomrule
    \end{tabular}
    
    \begin{tablenotes}
       \footnotesize
      \item The dataset was divided into training/validation/testing sets in a 7:1:2 ratio. The mean and standard deviation of performances run with three random seed are reported. The baselines' performances are taken from CASTER~\cite{huang2020caster}. 
     \end{tablenotes}
    \end{threeparttable}
\end{table*}
In MPG, we assigned a segmentation embedding to every node and every edge indicating which subgraph it belongs to (details are referred to Section \ref{sssec:input_repr}).  This deliberate design endows the model with the capability of taking simultaneous two graphs inputs. In this way, our MPG can be conveniently applied in some tasks with graph pair input, such as commonly used drug-drug interaction (DDI). DDI describes the interactions that one drug may affect others' activities when multiple drugs are administered simultaneously~\cite{rodrigues2019drug}. As the interaction among drugs could trigger an unexpected negative or positive impact on the therapeutic outcomes, characterizing DDI is extremely important for improving drug consumption safety and efficacy.
To demonstrate the effectiveness of MPG on DDI prediction, we compared our framework against the recently proposed algorithms on two real-world datasets---BIOSNAP\cite{biosnapnets} and TWOSIDES\cite{tatonetti2012data} (Details about both datasets are referred to SI).
To ensure a fair comparison, we followed the identical experimental procedure of two best approaches---CASTER~\cite{huang2020caster} and DDI-PULearn~\cite{zheng2019ddi}, on above two datasets, respectively. The DDI prediction tasks are formalized as a binary classification problem that aims to identify an interaction between two drugs. The classification results are reported in Table \ref{DDI_pred_figure} and Figure \ref{fig:twosides}.

Table \ref{DDI_pred_figure} and Figure \ref{fig:twosides} show that MPG significantly outperforms the previous best methods (CASTER and DDI-PULearn) on both two datasets by a large margin (7\% and 9\% improvements in terms of F1 score, respectively). CASTER take SMILES~\cite{weininger1988smiles}  sub-strings as inputs to represent molecular sub-structure. Compared to SMILES, a hydrogen-depleted molecular graph is more suitable and effective to represent molecules' structural information~\cite{wu2018moleculenet}. DDI-PULearn~\cite{zheng2019ddi} collected various drug properties to calculate the drug-drug similarities as input representation, including drug chemical substructures, drug targets, side-effects, and drug indications. In contrast, our MPG only takes the molecular structure as inputs, and we observed that MPG still yielded significantly better performance than DDI-PULearn. These results demonstrate the prediction superior performance of MPG on DDI prediction.

\begin{figure}[ht]
\centering
\includegraphics[width=0.7\linewidth]
{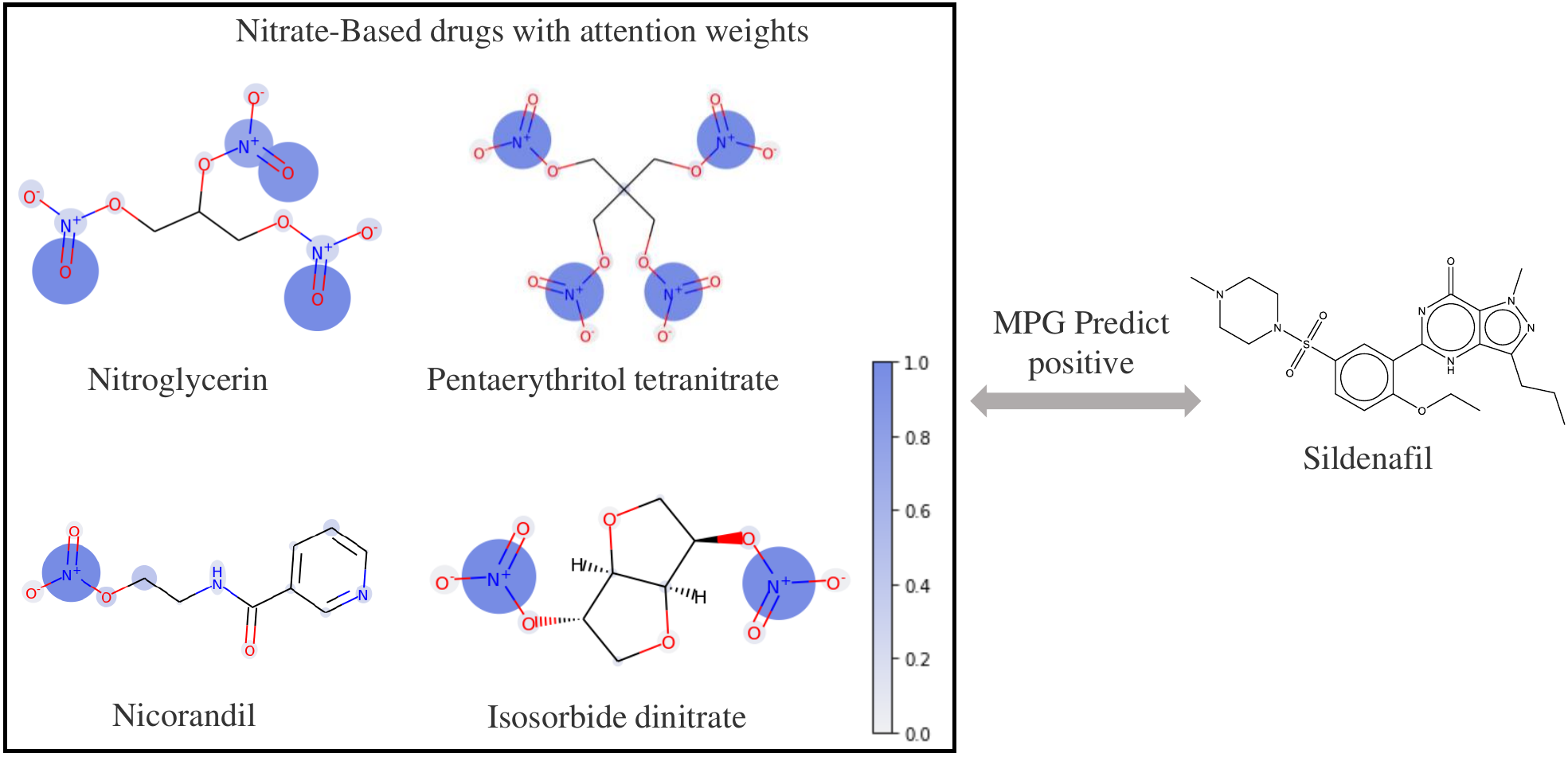}
\caption{MPG provides explainability for DDI prediction. The left part of the figure illustrates nitrate-based drugs with each atom colored by the attention weight. Note that our MPG always pays more attention to the nitrate group.}
\label{fig:ddi}
\end{figure}

Furthermore, MPG can generate an interpretable prediction. Given an input drug pair, MPG assigns an attention weight to each atom in molecules, indicating the importance of the interaction.  We chose the interaction between Sildenafil and other Nitrate-Based drugs as a case study.  Sildenafil, a PDE5 inhibitor, is developed as an effective treatment for pulmonary hypertension and erectile dysfunction. Because Nitrate-Based drugs and Sildenafil increase cGMP (nitrates increase cGMP formation and Sildenafil decrease cGMP breakdown ), it could lead to intense drops with blood pressure and even heart attack when used in combination. Thus, we would test if our MPG can pay more attention to the nitrate group when it predicts the interaction between Sildenafil and other nitrate-Based drugs. Specifically, we extracted and normalized the atom's attention weights to the collection nodes from the last layer of \ourmodel. After visualizing the attention weights, we observed that there always exists high attention weights on the nitrate group (Figure \ref{fig:ddi}). This suggests that MPG could 
leverage sparse and reasonable information of molecules to generate DDI prediction.

\subsection{MPG boosts the performance of drug-target interaction prediction}
\begin{figure}[ht]
\centering
\includegraphics[width=1\linewidth]
{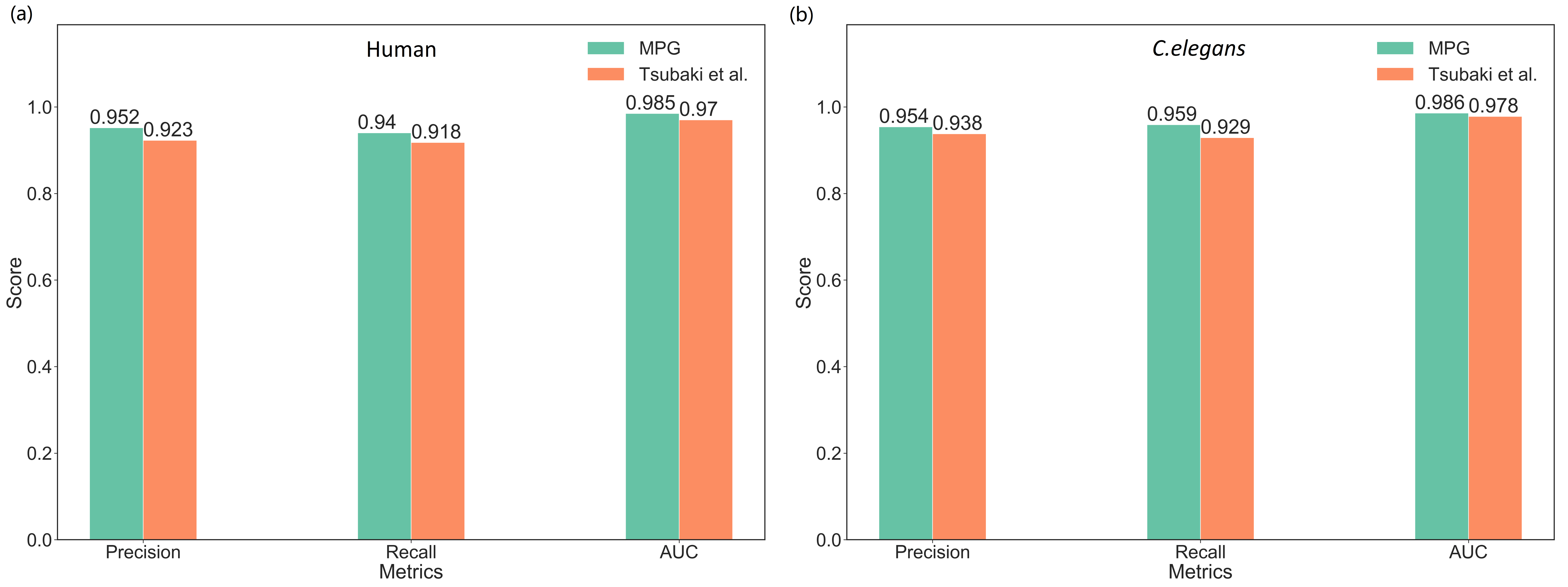}
\caption{Classification performance for DTI prediction on Human (a) and \textit{C.elegans} (b) datasets.}
\label{fig:DTI}
\end{figure}

As experiments above show that MPG achieves impressive performance on ligand-based CADD tasks, we further explored MPG's capacity on structure-based CADD. Structure-based CADD aims to identify the interaction between the compound and target protein for drug discovery. Various deep learning methods have been developed and achieved excellent performance for drug-target interaction (DTI) prediction~\cite{mousavian2014drug,chen2018machine,wen2017deep}. Generally, the deep learning algorithms for DTI prediction comprise of a compound encoder and a protein encoder. Recently, Tsubaki et al.~\cite{Tsubaki2019} proposed a framework that employed GNN and CNN for compound and protein sequence encoding, respectively, and leveraged an attention mechanism to integrate information of compound and protein to predict the DTI, which significantly outperformed existing methods. Here, we adapted their framework and replaced their compound encoder with our MPG to evaluate its effectiveness on DTI prediction (as shown in Figure S3). We followed the same experimental procedure as Tsubaki et al.~\cite{Tsubaki2019} to ensure a fair comparison on two widely used datasets---Human and\textit{ C.elegans} datasets. Figure \ref{fig:DTI} shows that our model outperforms Tsubaki's model on both two datasets, re-confirming MPG's powerful capacity for modeling molecules. 

\subsection{Ablation studies}

\subsubsection{Pre-train and No pre-train}
To verify the necessity of pre-training in MPG, we compared the performances of pre-trained and non-pre-trained \ourmodel ~on molecular properties prediction tasks, both of which have the identical hyper-parameter setting. Table \ref{tab:ablation} shows that, compared with the pre-trained \ourmodel, the \ourmodel~ without pre-training demonstrates significant decreases in classification AUC-ROC score, and increases in RMSE of regression tasks, which confirmed that our self-supervised strategies could provide a favorable initialization for the model and improve the performance of downstream tasks. Notably, the small datasets, including BBBP, SIDER, ClinTox, BACE, and FreeSolv, give a greater performance gain through pre-training, demonstrating the effectiveness and generalizability of the self-supervised pre-training for tasks with insufficient labeled molecules.

\begin{table*}[htbp]
  \centering
  \caption{Effect of pre-training strategies.}

   \resizebox{1.0\textwidth}{!}{
     \begin{threeparttable} 
    \begin{tabular}{c|c|c|cccccc|ccc}
    \hline

    \multicolumn{3}{c|}{\textbf{Datasets}} & \textbf{BBBP} & \textbf{Tox21} & \textbf{ToxCast} & \textbf{SIDER} & \textbf{ClinTox} & \textbf{BACE} & \textbf{FreeSolv} & \textbf{ESOL} & \textbf{Lipo} \\
    \hline
    \multicolumn{3}{c|}{\textbf{Molecules}} & 2039  & 7831  & 8575  & 1427  & 1478  & 1513  & 642   & 1128  & 4200 \\
    \hline
         & AttrMasking &  PSD &  \multicolumn{6}{c|}{\textbf{Classification (AUC-ROC \%)}} & \multicolumn{3}{c}{\textbf{Regression (RMSE)}} \\
    \hline

      MPG(no pre-train)  & - & -  & 89.2(0.8) & 80.1(1.2) & 69.9(1.6) & 58.5(1.5) & \textcolor[rgb]{ .051,  0,  .082}{92.4 (3.4)} & 86.8(1.4) & 1.967(0.556) & 0.896(0.145) & 0.628(0.062) \\
    MPG(node-level)  & \checkmark & -& 90.2(1.5) & 81.9(0.8) & 72.6(0.7) & 61.1(0.8) & 93.5(2.3) & 87.7(1.4) & 1.829(0.172) & 0.835(0.192) & 0.710(0.049) \\
    
    MPG(graph-level)  &-  &\checkmark  & 91.1(0.8) & 83.4(1.0) & 72.2(1.0) & 62.2(0.7) & \textcolor[rgb]{ .051,  0,  .082}{95.1 (1.5)} & 88.4(0.8) & 1.464(0.196) & 0.814(0.067) & 0.608(0.021) \\
  MPG  & \checkmark &\checkmark & \textbf{92.2(1.2)} & \textbf{83.7(1.9)} & \textbf{74.8(0.5)} & \textbf{65.8(1.2)} & \textbf{96.3(2.8)} & \textbf{92.0(1.3)} & \textbf{1.269(0.192)} & \textbf{0.802(0.043)} & \textbf{0.576(0.029)} \\
    \hline
    \end{tabular}%
 \end{threeparttable} 
    }

  \label{tab:ablation}%
\end{table*}%

\subsubsection{The effect of PSD strategy}
In the pre-training process of MPG, we employed AtrrMasking and PSD strategies to jointly pre-train \ourmodel. To investigate the contributions of these two strategies, we pre-trained our model with AtrrMasking or with PSD separately to compare their performance on downstream classification tasks. These self-supervised strategies follow the same hyper-parameter setting. Table \ref{tab:ablation} shows that both strategies can improve the average AUC-ROC score compared with no pre-training. Meantime, our PSD strategies outperform AtrrMasking on 8 out of 9 data sets, which indicates the importance and superiority of graph-level self-supervised learning for molecular properties prediction. It should be noted that combining these two strategies for pre-training yields a greater improvement than pre-training through either strategies.

%% file: discuss.tex
\section{Discussion}
\textbf{Molecular Representations.}
Molecular representations can be generally categorized into handcrafted representations and learned representations. 
Fingerprint~\cite{heinonen2012metabolite} and SMILES~\cite{weininger1988smiles} are two widely used handcrafted representations.
The most common type of fingerprint is a series of binary digits (bits) representing the presence or absence of particular substructures in the molecule. Although molecular fingerprint features in its flexibility and ease of computation for reaction prediction~\cite{segler17}, it also gives rise to several issues, including bit collisions and vector sparsity. Besides, molecules can be encoded as SMILES~\cite{weininger1988smiles} in the format of single-line text. Nevertheless, a key weakness in representing molecules using text sequences is its fragility of the representation, since small changes in the text sequence can lead to a large change in the molecular structure. 
Compared to the handcrafted representations, the learned molecular representation by deep learning has better generalization and higher expressive power, but it usually lacks explainability. That is, we have no idea about how the representation generates and what the representation stands for. This study makes a attempt to investigate the explainability of molecular representaion, and found that our MPG can capture some chemistry knowledge. Nevertheless, we still know too little about it compared to what it is all about. Further analysis both theoretically and empirically are desired to better understand when/why/how pre-training for GNNs can work.

\textbf{Self-supervised strategies.} 
Self-supervised strategies have crucial impact on performance of pre-trained model.
Current self-supervised strategies for pre-training GNNs suffer from either high computation complexity or falling into node-level learning, which are time-consuming and ineffective when applied in large-scale molecule pre-training. Here,  we applied three main principles for designing an appropriate self-supervised strategy to pre-train on molecule---computation-friendly, architecture-free, node and graph-level learning. First, our strategy is computation-friendly that enables the model to pre-train on large scale data to encode more information.
Second, the strategy is independent of the model, as we may evaluate different models to select the optimal architecture. Last, our strategy can pre-train the model at both node-level and graph-level to encode more information of structural characteristic.
This work serves as an important first step towards the graph-level self-supervised learning on large scale molecule data. Although we focused on molecular representation for drug discovery, the approach presented in this work is more general, and can be adapted to any graph representation learning for other areas, such as social networks.

%% file: method.tex
\section{Methods}
\label{sec:method}
In this section, we firstly introduce the essential components of  \ourmodel, then we describe the self-supervised strategy---PSD in detail.

\subsection{\ourmodel ~Model}
\label{sec:model}
\ourmodel ~consists of three key components: graph attention module, feed forward network, and vertex update function. We will elaborate on these three components in the following.
\subsubsection{Neighbor Attention Module}
The input to \textit{neighbor attention module} at time step $t$ is a set of atom representation $\bm x=\{x_1^{t-1},\cdots,\\x_N^{t-1}\},x_i^{t-1}\in \mathbb{R}^d$ and a set of bond representation $\bm e=\{\cdots,e_{i,j},\cdots\},e_{i,j}\in \mathbb{R}^d$. The module captures the interaction information between the atom and its neighbors (including its neighbor atoms and neighbor edges) to produce a message representation for each node $\bm m=\{m_1^t,\cdots,m_N^t\},$
$m_i^t\in \mathbb{R}^d$.

For each atom $i$, the \textit{neighbor attention module} first adds atom $i$'s neighbor atom representation  $x_j^t$ with the edge $e_{i,j}$ between them to represent the neighbor information $I_{j}^t$, that is:
\begin{equation}
    I_{j}^t=x_j^{t-1}+e_{i,j}
\end{equation}
Given the neighbor information and atom representation, the module performs scaled dot-product attention~\cite{vaswani2017attention} on the atoms---a shared attention mechanism computes the attention score. Formally,  we firstly map the node $x_i^t$ into query $Q_i^t$, and map its neighbor information $I_j^t$ into key $K_j^t$ and value $V_j^t$ respectively, computed by

\begin{align}
Q_i^t &=  \bm W_qx_i^{t-1} \\
K_j^t &=  \bm W_kI_j^t\\
V_j^t &=  \bm W_vI_j^t
\end{align}
where $\bm W_k$, $\bm W_q$ and $\bm W_v$ are the learnable weights matrices shared across all nodes, the dimension of $Q_i^t$ and $K_j^t$ is $d_k$, and the dimension of $V_j^t$ is $d$. We compute the dot products of the query and key to indicate the importance of neighbor information to node $i$. To avoid that the dot products grow large in magnitude, we scale the dot products by $\frac{1}{\sqrt{d_k}}$. That is
\begin{equation}
    s_{i,j}^t = \frac{Q_i^tK_j^t{^T}}{\sqrt{d_k}}
\end{equation}
To make coefficients easily comparable across different nodes, we then normalize them across all choices of $j$ using the softmax function:
\begin{equation}
    a_{i,j}^t = \text{softmax}(s_{i,j}^t) = \frac{e^{s_{i,j}^t}}{\sum_{j\in \mathcal{N}_i} e^{s_{i,j}^t}} ,
\end{equation}
where $\mathcal{N}_i$ stands for the neighbors of node $i$.

Once obtained, the normalized attention coefficients together with neighbor values $V_j$ are used to apply weighted summation operation, to derive the message representation $m_i^t$ for every node:
\begin{equation}
\label{eq:weightsum}
m_i^t = \sum_{j\in \mathcal{N}_i} a_{i,j}^t V_j^t,
\end{equation}

The \textit{neighbor attention module} also employs multi-head attention to stabilize the learning process of self-attention, that is, $K$ independent attention mechanisms execute the transformation of Equation~\ref{eq:weightsum}, and then their features are concatenated, fed into a linear transformation, resulting in the following output representation:
\begin{equation}
m_i^t = \bm W_m||_k^K \sum_{j\in \mathcal{N}_i} a_{i,j}^{t,k} V_j^{t,k},
\end{equation}
where $||$ represents concatenation, $a_{i,j}^{t,k}$ is the normalized attention coefficients computed by the $k$-th attention mechanism, $V_j^{t,k}$ is the corresponding neighbor value, $W_m$ is a learnable weight matrix shared across all nodes. 
\subsubsection{Feed-forward Network}
To extract a deep representation of the message and increase the expression power of model, we feed the message representation extracted by \textit{neighbor attention module} into a fully connected feed-forward network. This network consists of two linear transformations with a Gaussian Error Linear Unit (GELU)~\cite{hendrycks2016gaussian} activation in between.
\begin{equation}
    m_i^t=\bm W_2\sigma(\bm W_1m_i^t+b_1)+b_2
\end{equation}
where $W_1\in \mathbb{R}^{d_{ff} \times d}$ and $W_2 \in \mathbb{R}^{d \times d_{ff}}$ are learnable weight matrix, $\sigma$ is GELU activation function. In our experiments, the dimension $d_{ff}$ is four times of $d$, that is 3072 ($d=768$).

\subsubsection{Vertex Update Function}
Based on the properly represented neighbor message $m_i^t$, our model \ourmodel ~employs a GRU network~~\cite{cho2014learning} to update the atom's representation $x_i^t$, computed by

\begin{align}
    r^{t}_i&=sigmoid(\bm W_{mr}m_i^t+b_{mr}+\bm W_{xr}h^{t-1}_i+b_{hr})\\
    u^{t}_i&=sigmoid(\bm W_{mu}m_i^t+b_{mu}+\bm W_{xu}h^{t-1}_i+b_{hu})\\
    x^{t}_i&=tanh(\bm W_{in}m_i^t+b_{in}+r^{t}_i*(\bm W_{hn}h^{t-1}_i+b_{hn}))\\
    h^{t}_i&=(1-u^{t}_i)*x^{t-1}_i+u^{t}_i*x_i^t
\end{align}
where $h^t_i$ is the hidden state of atom $i$ in GRU at time $t$, $h_i^{t-1}$ is the hidden state at time $t-1$, the initial hidden state $h_i^0$ is the atom representation $x_i^0$, and $r_i^t$ and $u_i^t$ are the reset and update gate, respectively. $*$ is the Hadamard product.

\subsection{PSD Strategy}
Simply, PSD task is designed to discriminate whether two subgraphs come from the same source, in other words, are homologous. As shown in Figure \ref{fig:overview}, the graph is firstly decomposed into two subgraphs, one of these two subgraphs has a 0.5 chance to be replaced by a subgraph disconnected from another graph which constitutes the negative sample, otherwise the positive sample. 
We employed the cross-entropy loss function instead of NCE~\cite{mnih2013learning} for simple computation to optimize the parameters of the network as follows:
\begin{equation}
    L = -\sum_{i=1}^m y\log(p)+(1-y)\log(1-p)
\end{equation}
where $m$ is the number of samples. After pre-trained, the collection node embedding can be regarded as a graph-level representation for the graph and used for downstream tasks. In addition, graph representation can also be obtained by averaging the nodes' embeddings or other global graph pooling methods.

In the following sections, we describe the important components of PSD in detail.

\subsubsection{Graph Decomposition and Negative Sampling}
\begin{figure}[ht]
\centering
\includegraphics[width=0.45\linewidth]
{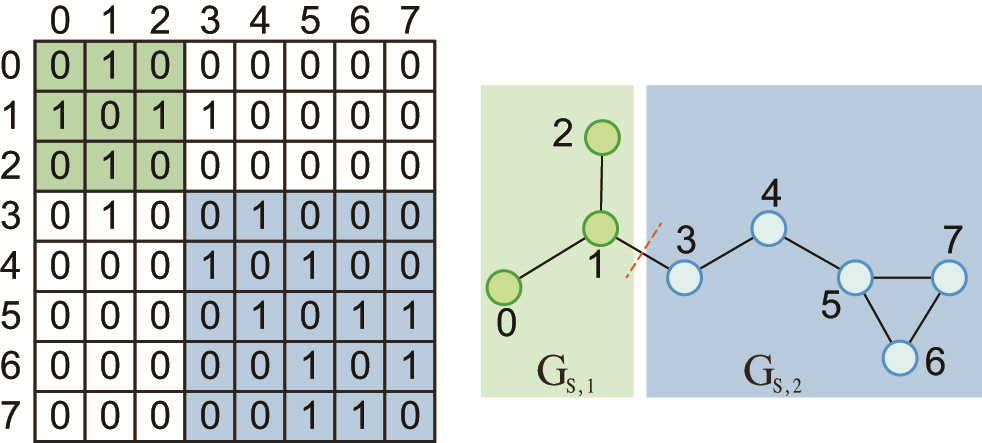}
\caption{The graph decomposition sample. The left sub-figure is the adjacency matrix of the graph in the right sub-figure, where the green and blue represent the decomposed two subgraphs.}
\label{fig:decompo}
\end{figure}

We decompose the graph into two subgraphs to generate the subgraph pairs, served as the positive sample, and replace one of the subgraphs to produce the negative sample. As the example shown in Figure \ref{fig:decompo}, given a graph $G=(V,E)$ where $V$ represents nodes and $E$ represents edges. A sampled node $v_3$ is employed as the border node to separate $G$ into two subgraphs $G_{s,1}$ and $G_{s,2}$, where $G_{s,1}$ contains nodes $\{v_0,v_1,v_2\}$ and $G_{s,2}$ contains nodes $\{v_3,v_4,\cdots,v_7\}$. The edges in these two subgraphs correspond to the top-left sub-matrix and bottom-right sub-matrix of the adjacency matrix respectively. In order to produce subgraphs with balanced and various size, the border node index is randomly sampled in the range of 1/3 to 2/3 of the total number of nodes.
For negative sampling, we randomly sample another graph in the dataset and separate it into two subgraphs using the above method, and $G_{s,2}$ is replaced with one of these two subgraphs to generate a negative sample.
How negative samples are generated can have a large impact on the quality of the learned embeddings. It may drive the model to identify whether the two graphs are homologous or estimate whether the two graphs can be combined into a valid graph. In this way, the model can learn the valuable graph-level features of graphs from the nodes and edges which is essential for the downstream tasks.
\subsubsection{Virtual Collection Node}
The subgraph pair obtained via the above approach are two independent graphs without any connection. 
We concatenate these two subgraphs into a single whole graph, and introduce a virtual collection node to derive the global graph-level representation by aggregating every node information. The collection node is linked with all the other nodes by virtual directed edges, pointing from the other nodes to the collection node. During the message passing process of GNN, the collection node learns its representation from all the other nodes but does not affect the feature update procedure of them. Consequently, the collection node's feature can grasp the global representation of the subgraphs pair and be fed into a feed-forward neural network for the final prediction.

\subsubsection{Input Representation}
\label{sssec:input_repr}
\begin{figure}[ht]
\centering
\includegraphics[width=0.35\linewidth]
{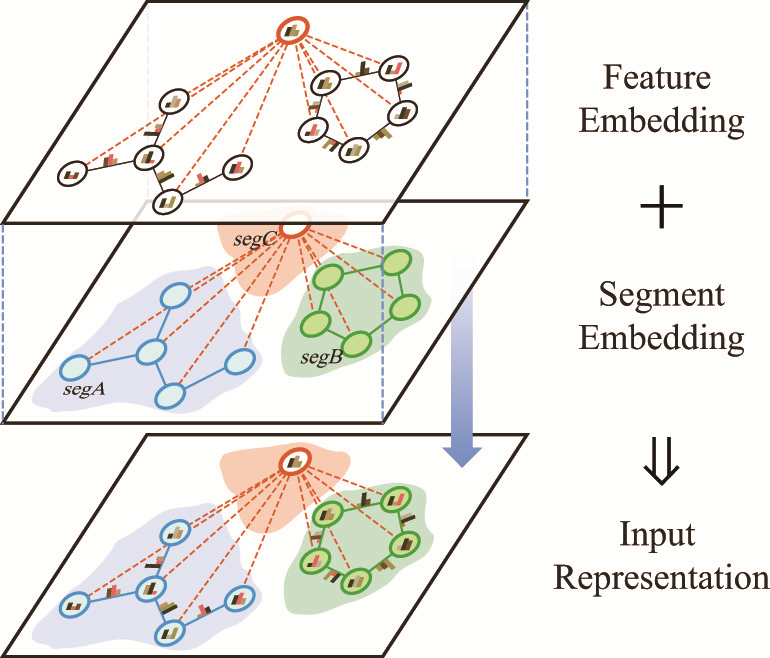}
\caption{The input representation of graph data is constructed by summing two parts: feature embedding and segment embedding. (a) Feature embedding: a set of node and edge features go through the embedding transformation to describe a graph. (b) Segment embedding: a learned segmentation embedding to every node and every edge indicating which subgraph it belongs to,  different colors represent different segmentation.}
\label{fig:representation}
\end{figure}

As shown in Figure \ref{fig:representation}, the input representation consists of two parts: feature embedding and segment embedding. A graph is generally described by a set of nodes features and edges features as shown in Table S1.  Besides the feature embedding, we add a learned segmentation embedding to every node and every edge indicating which subgraph it belongs to. The final input representation is constructed by summing the segment embedding and feature embedding. In this way, the model could distinguish the nodes and edges from different segments, thus enables simultaneous input of two graphs. 

%% file: MPG.bbl
\begin{thebibliography}{10}
\expandafter\ifx\csname url\endcsname\relax
  \def\url#1{\texttt{#1}}\fi
\expandafter\ifx\csname urlprefix\endcsname\relax\def\urlprefix{URL }\fi
\providecommand{\bibinfo}[2]{#2}
\providecommand{\eprint}[2][]{\url{#2}}

\bibitem{hill2012drug}
\bibinfo{author}{Hill, R.~G.}
\newblock \emph{\bibinfo{title}{Drug discovery and development-E-book:
  technology in transition}} (\bibinfo{publisher}{Elsevier Health Sciences},
  \bibinfo{year}{2012}).

\bibitem{chan2019advancing}
\bibinfo{author}{Chan, H.~S.}, \bibinfo{author}{Shan, H.},
  \bibinfo{author}{Dahoun, T.}, \bibinfo{author}{Vogel, H.} \&
  \bibinfo{author}{Yuan, S.}
\newblock \bibinfo{title}{Advancing drug discovery via artificial
  intelligence}.
\newblock \emph{\bibinfo{journal}{Trends in pharmacological sciences}}
  \textbf{\bibinfo{volume}{40}}, \bibinfo{pages}{592--604}
  (\bibinfo{year}{2019}).

\bibitem{sliwoski2014computational}
\bibinfo{author}{Sliwoski, G.}, \bibinfo{author}{Kothiwale, S.},
  \bibinfo{author}{Meiler, J.} \& \bibinfo{author}{Lowe, E.~W.}
\newblock \bibinfo{title}{Computational methods in drug discovery}.
\newblock \emph{\bibinfo{journal}{Pharmacological reviews}}
  \textbf{\bibinfo{volume}{66}}, \bibinfo{pages}{334--395}
  (\bibinfo{year}{2014}).

\bibitem{kapetanovic2008computer}
\bibinfo{author}{Kapetanovic, I.}
\newblock \bibinfo{title}{Computer-aided drug discovery and development
  (caddd): in silico-chemico-biological approach}.
\newblock \emph{\bibinfo{journal}{Chemico-biological interactions}}
  \textbf{\bibinfo{volume}{171}}, \bibinfo{pages}{165--176}
  (\bibinfo{year}{2008}).

\bibitem{ghasemi2018neural}
\bibinfo{author}{Ghasemi, F.}, \bibinfo{author}{Mehridehnavi, A.},
  \bibinfo{author}{Perez-Garrido, A.} \& \bibinfo{author}{Perez-Sanchez, H.}
\newblock \bibinfo{title}{Neural network and deep-learning algorithms used in
  qsar studies: merits and drawbacks}.
\newblock \emph{\bibinfo{journal}{Drug Discov. Today}}
  \textbf{\bibinfo{volume}{23}}, \bibinfo{pages}{1784--1790}
  (\bibinfo{year}{2018}).

\bibitem{ryu2018deep}
\bibinfo{author}{Ryu, J.~Y.}, \bibinfo{author}{Kim, H.~U.} \&
  \bibinfo{author}{Lee, S.~Y.}
\newblock \bibinfo{title}{Deep learning improves prediction of drug--drug and
  drug--food interactions}.
\newblock \emph{\bibinfo{journal}{Proceedings of the National Academy of
  Sciences}} \textbf{\bibinfo{volume}{115}}, \bibinfo{pages}{E4304--E4311}
  (\bibinfo{year}{2018}).

\bibitem{abbasi2020deep}
\bibinfo{author}{Abbasi, K.}, \bibinfo{author}{Razzaghi, P.},
  \bibinfo{author}{Poso, A.}, \bibinfo{author}{Ghanbari-Ara, S.} \&
  \bibinfo{author}{Masoudi-Nejad, A.}
\newblock \bibinfo{title}{Deep learning in drug target interaction prediction:
  Current and future perspective}.
\newblock \emph{\bibinfo{journal}{Current Medicinal Chemistry}}
  (\bibinfo{year}{2020}).

\bibitem{d2020machine}
\bibinfo{author}{D’Souza, S.}, \bibinfo{author}{Prema, K.} \&
  \bibinfo{author}{Balaji, S.}
\newblock \bibinfo{title}{Machine learning models for drug--target
  interactions: current knowledge and future directions}.
\newblock \emph{\bibinfo{journal}{Drug Discovery Today}}
  \textbf{\bibinfo{volume}{25}}, \bibinfo{pages}{748--756}
  (\bibinfo{year}{2020}).

\bibitem{yang2019analyzing}
\bibinfo{author}{Yang, K.} \emph{et~al.}
\newblock \bibinfo{title}{Analyzing learned molecular representations for
  property prediction}.
\newblock \emph{\bibinfo{journal}{Journal of chemical information and
  modeling}} \textbf{\bibinfo{volume}{59}}, \bibinfo{pages}{3370--3388}
  (\bibinfo{year}{2019}).

\bibitem{xue2000molecular}
\bibinfo{author}{Xue, L.} \& \bibinfo{author}{Bajorath, J.}
\newblock \bibinfo{title}{Molecular descriptors in chemoinformatics,
  computational combinatorial chemistry, and virtual screening}.
\newblock \emph{\bibinfo{journal}{Combinatorial chemistry \& high throughput
  screening}} \textbf{\bibinfo{volume}{3}}, \bibinfo{pages}{363--372}
  (\bibinfo{year}{2000}).

\bibitem{gilmer2017neural}
\bibinfo{author}{Gilmer, J.}, \bibinfo{author}{Schoenholz, S.~S.},
  \bibinfo{author}{Riley, P.}, \bibinfo{author}{Vinyals, O.} \&
  \bibinfo{author}{Dahl, G.~E.}
\newblock \bibinfo{title}{Neural message passing for quantum chemistry}.
\newblock \emph{\bibinfo{journal}{international conference on machine
  learning}} \bibinfo{pages}{1263--1272} (\bibinfo{year}{2017}).

\bibitem{kipf2016semi}
\bibinfo{author}{Kipf, T.~N.} \& \bibinfo{author}{Welling, M.}
\newblock \bibinfo{title}{Semi-supervised classification with graph
  convolutional networks}.
\newblock \emph{\bibinfo{journal}{arXiv preprint arXiv:1609.02907}}
  (\bibinfo{year}{2016}).

\bibitem{velivckovic2017graph}
\bibinfo{author}{Veli{\v{c}}kovi{\'c}, P.} \emph{et~al.}
\newblock \emph{\bibinfo{title}{Graph attention networks}}
  (\bibinfo{publisher}{ICLR}, \bibinfo{year}{2018}).

\bibitem{hamilton2017inductive}
\bibinfo{author}{Hamilton, W.~L.}, \bibinfo{author}{Ying, R.} \&
  \bibinfo{author}{Leskovec, J.}
\newblock \bibinfo{title}{Inductive representation learning on large graphs}.
\newblock \bibinfo{pages}{1025--1035} (\bibinfo{year}{2017}).

\bibitem{Wu2018}
\bibinfo{author}{Wu, Z.} \emph{et~al.}
\newblock \bibinfo{title}{{MoleculeNet: A benchmark for molecular machine
  learning}}.
\newblock \emph{\bibinfo{journal}{Chemical Science}}
  \textbf{\bibinfo{volume}{9}}, \bibinfo{pages}{513--530}
  (\bibinfo{year}{2018}).
\newblock \eprint{1703.00564}.

\bibitem{hu2019strategies}
\bibinfo{author}{Hu, W.} \emph{et~al.}
\newblock \bibinfo{title}{Strategies for pre-training graph neural networks}.
\newblock In \emph{\bibinfo{booktitle}{International Conference on Learning
  Representations}} (\bibinfo{year}{2019}).

\bibitem{rong2020grover}
\bibinfo{author}{Rong, Y.} \emph{et~al.}
\newblock \bibinfo{title}{Self-supervised graph transformer on large-scale
  molecular data}.
\newblock \emph{\bibinfo{journal}{Advances in Neural Information Processing
  Systems}} \textbf{\bibinfo{volume}{33}} (\bibinfo{year}{2020}).

\bibitem{liu2020self}
\bibinfo{author}{Liu, X.} \emph{et~al.}
\newblock \bibinfo{title}{Self-supervised learning: Generative or contrastive}.
\newblock \emph{\bibinfo{journal}{arXiv}} \bibinfo{pages}{arXiv--2006}
  (\bibinfo{year}{2020}).

\bibitem{krizhevsky2012imagenet}
\bibinfo{author}{Krizhevsky, A.}, \bibinfo{author}{Sutskever, I.} \&
  \bibinfo{author}{Hinton, G.~E.}
\newblock \bibinfo{title}{Imagenet classification with deep convolutional
  neural networks}.
\newblock \bibinfo{pages}{1097--1105} (\bibinfo{year}{2012}).

\bibitem{he2020momentum}
\bibinfo{author}{He, K.}, \bibinfo{author}{Fan, H.}, \bibinfo{author}{Wu, Y.},
  \bibinfo{author}{Xie, S.} \& \bibinfo{author}{Girshick, R.}
\newblock \bibinfo{title}{Momentum contrast for unsupervised visual
  representation learning}.
\newblock In \emph{\bibinfo{booktitle}{Proceedings of the IEEE/CVF Conference
  on Computer Vision and Pattern Recognition}}, \bibinfo{pages}{9729--9738}
  (\bibinfo{year}{2020}).

\bibitem{devlin2019bert}
\bibinfo{author}{Devlin, J.}, \bibinfo{author}{Chang, M.-W.},
  \bibinfo{author}{Lee, K.} \& \bibinfo{author}{Toutanova, K.}
\newblock \bibinfo{title}{Bert: Pre-training of deep bidirectional transformers
  for language understanding}.
\newblock In \emph{\bibinfo{booktitle}{Proceedings of the 2019 Conference of
  the North American Chapter of the Association for Computational Linguistics:
  Human Language Technologies, Volume 1 (Long and Short Papers)}},
  \bibinfo{pages}{4171--4186} (\bibinfo{year}{2019}).

\bibitem{weininger1988smiles}
\bibinfo{author}{Weininger, D.}
\newblock \bibinfo{title}{Smiles, a chemical language and information system.
  1. introduction to methodology and encoding rules}.
\newblock \emph{\bibinfo{journal}{Journal of Chemical Information and Computer
  Sciences}} \textbf{\bibinfo{volume}{28}}, \bibinfo{pages}{31--36}
  (\bibinfo{year}{1988}).

\bibitem{honda2019smiles}
\bibinfo{author}{Honda, S.}, \bibinfo{author}{Shi, S.} \&
  \bibinfo{author}{Ueda, H.~R.}
\newblock \bibinfo{title}{Smiles transformer: Pre-trained molecular fingerprint
  for low data drug discovery}.
\newblock \emph{\bibinfo{journal}{arXiv preprint arXiv:1911.04738}}
  (\bibinfo{year}{2019}).

\bibitem{pesciullesi2020transfer}
\bibinfo{author}{Pesciullesi, G.}, \bibinfo{author}{Schwaller, P.},
  \bibinfo{author}{Laino, T.} \& \bibinfo{author}{Reymond, J.-L.}
\newblock \bibinfo{title}{Transfer learning enables the molecular transformer
  to predict regio-and stereoselective reactions on carbohydrates}.
\newblock \emph{\bibinfo{journal}{Nature communications}}
  \textbf{\bibinfo{volume}{11}}, \bibinfo{pages}{1--8} (\bibinfo{year}{2020}).

\bibitem{wang2019smiles}
\bibinfo{author}{Wang, S.}, \bibinfo{author}{Guo, Y.}, \bibinfo{author}{Wang,
  Y.}, \bibinfo{author}{Sun, H.} \& \bibinfo{author}{Huang, J.}
\newblock \bibinfo{title}{Smiles-bert: large scale unsupervised pre-training
  for molecular property prediction}.
\newblock In \emph{\bibinfo{booktitle}{Proceedings of the 10th ACM
  International Conference on Bioinformatics, Computational Biology and Health
  Informatics}}, \bibinfo{pages}{429--436} (\bibinfo{year}{2019}).

\bibitem{chithrananda2020chemberta}
\bibinfo{author}{Chithrananda, S.}, \bibinfo{author}{Grand, G.} \&
  \bibinfo{author}{Ramsundar, B.}
\newblock \bibinfo{title}{Chemberta: Large-scale self-supervised pretraining
  for molecular property prediction}.
\newblock \emph{\bibinfo{journal}{arXiv preprint arXiv:2010.09885}}
  (\bibinfo{year}{2020}).

\bibitem{winter2019learning}
\bibinfo{author}{Winter, R.}, \bibinfo{author}{Montanari, F.},
  \bibinfo{author}{No{\'e}, F.} \& \bibinfo{author}{Clevert, D.-A.}
\newblock \bibinfo{title}{Learning continuous and data-driven molecular
  descriptors by translating equivalent chemical representations}.
\newblock \emph{\bibinfo{journal}{Chemical science}}
  \textbf{\bibinfo{volume}{10}}, \bibinfo{pages}{1692--1701}
  (\bibinfo{year}{2019}).

\bibitem{gomez2018automatic}
\bibinfo{author}{G{\'o}mez-Bombarelli, R.} \emph{et~al.}
\newblock \bibinfo{title}{Automatic chemical design using a data-driven
  continuous representation of molecules}.
\newblock \emph{\bibinfo{journal}{ACS central science}}
  \textbf{\bibinfo{volume}{4}}, \bibinfo{pages}{268--276}
  (\bibinfo{year}{2018}).

\bibitem{xu2017seq2seq}
\bibinfo{author}{Xu, Z.}, \bibinfo{author}{Wang, S.}, \bibinfo{author}{Zhu, F.}
  \& \bibinfo{author}{Huang, J.}
\newblock \bibinfo{title}{Seq2seq fingerprint: An unsupervised deep molecular
  embedding for drug discovery}.
\newblock In \emph{\bibinfo{booktitle}{Proceedings of the 8th ACM international
  conference on bioinformatics, computational biology, and health
  informatics}}, \bibinfo{pages}{285--294} (\bibinfo{year}{2017}).

\bibitem{liu2018ngram}
\bibinfo{author}{Liu, S.}, \bibinfo{author}{Demirel, M.~F.} \&
  \bibinfo{author}{Liang, Y.}
\newblock \bibinfo{title}{N-gram graph: Simple unsupervised representation for
  graphs, with applications to molecules} (\bibinfo{year}{2018}).
\newblock \eprint{1806.09206}.

\bibitem{chen2020simple}
\bibinfo{author}{Chen, T.}, \bibinfo{author}{Kornblith, S.},
  \bibinfo{author}{Norouzi, M.} \& \bibinfo{author}{Hinton, G.}
\newblock \bibinfo{title}{A simple framework for contrastive learning of visual
  representations}.
\newblock \emph{\bibinfo{journal}{arXiv preprint arXiv:2002.05709}}
  (\bibinfo{year}{2020}).

\bibitem{oord2018representation}
\bibinfo{author}{Oord, A. v.~d.}, \bibinfo{author}{Li, Y.} \&
  \bibinfo{author}{Vinyals, O.}
\newblock \bibinfo{title}{Representation learning with contrastive predictive
  coding}.
\newblock \emph{\bibinfo{journal}{arXiv preprint arXiv:1807.03748}}
  (\bibinfo{year}{2018}).

\bibitem{velivckovic2018deep}
\bibinfo{author}{Veli{\v{c}}kovi{\'c}, P.} \emph{et~al.}
\newblock \bibinfo{title}{Deep graph infomax} (\bibinfo{year}{2019}).

\bibitem{sun2019infograph}
\bibinfo{author}{Sun, F.-Y.}, \bibinfo{author}{Hoffmann, J.},
  \bibinfo{author}{Verma, V.} \& \bibinfo{author}{Tang, J.}
\newblock \bibinfo{title}{Infograph: Unsupervised and semi-supervised
  graph-level representation learning via mutual information maximization}.
\newblock \emph{\bibinfo{journal}{arXiv preprint arXiv:1908.01000}}
  (\bibinfo{year}{2019}).

\bibitem{qiu2020gcc}
\bibinfo{author}{Qiu, J.} \emph{et~al.}
\newblock \bibinfo{title}{Gcc: Graph contrastive coding for graph neural
  network pre-training}.
\newblock In \emph{\bibinfo{booktitle}{Proceedings of the 26th ACM SIGKDD
  International Conference on Knowledge Discovery \& Data Mining}},
  \bibinfo{pages}{1150--1160} (\bibinfo{year}{2020}).

\bibitem{cho2014learning}
\bibinfo{author}{Cho, K.} \emph{et~al.}
\newblock \bibinfo{title}{Learning phrase representations using rnn
  encoder-decoder for statistical machine translation}.
\newblock \emph{\bibinfo{journal}{arXiv preprint arXiv:1406.1078}}
  (\bibinfo{year}{2014}).

\bibitem{liu2020towards}
\bibinfo{author}{Liu, M.}, \bibinfo{author}{Gao, H.} \& \bibinfo{author}{Ji,
  S.}
\newblock \bibinfo{title}{Towards deeper graph neural networks}.
\newblock In \emph{\bibinfo{booktitle}{Proceedings of the 26th ACM SIGKDD
  International Conference on Knowledge Discovery \& Data Mining}},
  \bibinfo{pages}{338--348} (\bibinfo{year}{2020}).

\bibitem{li2019deepgcns}
\bibinfo{author}{Li, G.}, \bibinfo{author}{Muller, M.},
  \bibinfo{author}{Thabet, A.} \& \bibinfo{author}{Ghanem, B.}
\newblock \bibinfo{title}{Deepgcns: Can gcns go as deep as cnns?}
\newblock In \emph{\bibinfo{booktitle}{Proceedings of the IEEE International
  Conference on Computer Vision}}, \bibinfo{pages}{9267--9276}
  (\bibinfo{year}{2019}).

\bibitem{dbindex1979}
\bibinfo{author}{{Davies}, D.~L.} \& \bibinfo{author}{{Bouldin}, D.~W.}
\newblock \bibinfo{title}{A cluster separation measure}.
\newblock \emph{\bibinfo{journal}{IEEE Transactions on Pattern Analysis and
  Machine Intelligence}} \textbf{\bibinfo{volume}{PAMI-1}},
  \bibinfo{pages}{224--227} (\bibinfo{year}{1979}).

\bibitem{sterling2015}
\bibinfo{author}{Sterling, T.} \& \bibinfo{author}{Irwin, J.~J.}
\newblock \bibinfo{title}{Zinc 15 – ligand discovery for everyone}.
\newblock \emph{\bibinfo{journal}{Journal of Chemical Information and
  Modeling}} \textbf{\bibinfo{volume}{55}}, \bibinfo{pages}{2324--2337}
  (\bibinfo{year}{2015}).
\newblock \bibinfo{note}{PMID: 26479676}.

\bibitem{gaulton2011chembl}
\bibinfo{author}{Gaulton, A.} \emph{et~al.}
\newblock \bibinfo{title}{{C}h{EMBL}: a large-scale bioactivity database for
  drug discovery}.
\newblock \emph{\bibinfo{journal}{Nucleic {A}cids {R}esearch}}
  \textbf{\bibinfo{volume}{40}}, \bibinfo{pages}{D1100--D1107}
  (\bibinfo{year}{2011}).

\bibitem{mcinnes2018umap}
\bibinfo{author}{McInnes, L.}, \bibinfo{author}{Healy, J.} \&
  \bibinfo{author}{Melville, J.}
\newblock \bibinfo{title}{Umap: Uniform manifold approximation and projection
  for dimension reduction}.
\newblock \emph{\bibinfo{journal}{arXiv preprint arXiv:1802.03426}}
  (\bibinfo{year}{2018}).

\bibitem{bemis1996properties}
\bibinfo{author}{Bemis, G.~W.} \& \bibinfo{author}{Murcko, M.~A.}
\newblock \bibinfo{title}{The properties of known drugs. 1. molecular
  frameworks}.
\newblock \emph{\bibinfo{journal}{Journal of medicinal chemistry}}
  \textbf{\bibinfo{volume}{39}}, \bibinfo{pages}{2887--2893}
  (\bibinfo{year}{1996}).

\bibitem{hu2016computational}
\bibinfo{author}{Hu, Y.}, \bibinfo{author}{Stumpfe, D.} \&
  \bibinfo{author}{Bajorath, J.}
\newblock \bibinfo{title}{Computational exploration of molecular scaffolds in
  medicinal chemistry: Miniperspective}.
\newblock \emph{\bibinfo{journal}{Journal of medicinal chemistry}}
  \textbf{\bibinfo{volume}{59}}, \bibinfo{pages}{4062--4076}
  (\bibinfo{year}{2016}).

\bibitem{ramsundar2015massively}
\bibinfo{author}{Ramsundar, B.} \emph{et~al.}
\newblock \bibinfo{title}{Massively multitask networks for drug discovery}.
\newblock \emph{\bibinfo{journal}{arXiv preprint arXiv:1502.02072}}
  (\bibinfo{year}{2015}).

\bibitem{kearnes2016molecular}
\bibinfo{author}{Kearnes, S.}, \bibinfo{author}{McCloskey, K.},
  \bibinfo{author}{Berndl, M.}, \bibinfo{author}{Pande, V.} \&
  \bibinfo{author}{Riley, P.}
\newblock \bibinfo{title}{Molecular graph convolutions: moving beyond
  fingerprints}.
\newblock \emph{\bibinfo{journal}{Journal of computer-aided molecular design}}
  \textbf{\bibinfo{volume}{30}}, \bibinfo{pages}{595--608}
  (\bibinfo{year}{2016}).

\bibitem{schutt2017schnet}
\bibinfo{author}{Sch{\"u}tt, K.} \emph{et~al.}
\newblock \bibinfo{title}{Schnet: A continuous-filter convolutional neural
  network for modeling quantum interactions}.
\newblock In \emph{\bibinfo{booktitle}{Advances in neural information
  processing systems}}, \bibinfo{pages}{991--1001} (\bibinfo{year}{2017}).

\bibitem{lu2019molecular}
\bibinfo{author}{Lu, C.} \emph{et~al.}
\newblock \bibinfo{title}{Molecular property prediction: A multilevel quantum
  interactions modeling perspective}.
\newblock In \emph{\bibinfo{booktitle}{Proceedings of the AAAI Conference on
  Artificial Intelligence}}, vol.~\bibinfo{volume}{33},
  \bibinfo{pages}{1052--1060} (\bibinfo{year}{2019}).

\bibitem{xiong2019pushing}
\bibinfo{author}{Xiong, Z.} \emph{et~al.}
\newblock \bibinfo{title}{Pushing the boundaries of molecular representation
  for drug discovery with the graph attention mechanism}.
\newblock \emph{\bibinfo{journal}{Journal of Medicinal Chemistry}}
  (\bibinfo{year}{2019}).

\bibitem{liu2019n}
\bibinfo{author}{Liu, S.}, \bibinfo{author}{Demirel, M.~F.} \&
  \bibinfo{author}{Liang, Y.}
\newblock \bibinfo{title}{N-gram graph: Simple unsupervised representation for
  graphs, with applications to molecules}.
\newblock In \emph{\bibinfo{booktitle}{Advances in Neural Information
  Processing Systems}}, \bibinfo{pages}{8466--8478} (\bibinfo{year}{2019}).

\bibitem{cherkasov2014qsar}
\bibinfo{author}{Cherkasov, A.} \emph{et~al.}
\newblock \bibinfo{title}{Qsar modeling: where have you been? where are you
  going to?}
\newblock \emph{\bibinfo{journal}{Journal of medicinal chemistry}}
  \textbf{\bibinfo{volume}{57}}, \bibinfo{pages}{4977--5010}
  (\bibinfo{year}{2014}).

\bibitem{vilar2014similarity}
\bibinfo{author}{Vilar, S.} \emph{et~al.}
\newblock \bibinfo{title}{Similarity-based modeling in large-scale prediction
  of drug-drug interactions}.
\newblock \emph{\bibinfo{journal}{Nature protocols}}
  \textbf{\bibinfo{volume}{9}}, \bibinfo{pages}{2147} (\bibinfo{year}{2014}).

\bibitem{jaeger2018mol2vec}
\bibinfo{author}{Jaeger, S.}, \bibinfo{author}{Fulle, S.} \&
  \bibinfo{author}{Turk, S.}
\newblock \bibinfo{title}{Mol2vec: unsupervised machine learning approach with
  chemical intuition}.
\newblock \emph{\bibinfo{journal}{Journal of chemical information and
  modeling}} \textbf{\bibinfo{volume}{58}}, \bibinfo{pages}{27--35}
  (\bibinfo{year}{2018}).

\bibitem{huang2020caster}
\bibinfo{author}{Huang, K.}, \bibinfo{author}{Xiao, C.},
  \bibinfo{author}{Hoang, T.}, \bibinfo{author}{Glass, L.} \&
  \bibinfo{author}{Sun, J.}
\newblock \bibinfo{title}{Caster: Predicting drug interactions with chemical
  substructure representation}.
\newblock In \emph{\bibinfo{booktitle}{Proceedings of the AAAI Conference on
  Artificial Intelligence}}, vol.~\bibinfo{volume}{34},
  \bibinfo{pages}{702--709} (\bibinfo{year}{2020}).

\bibitem{rodrigues2019drug}
\bibinfo{author}{Rodrigues, A.~D.}
\newblock \emph{\bibinfo{title}{Drug-drug interactions}}
  (\bibinfo{publisher}{CRC Press}, \bibinfo{year}{2019}).

\bibitem{biosnapnets}
\bibinfo{author}{Marinka~Zitnik, S.~M., Rok~Sosi\v{c}} \&
  \bibinfo{author}{Leskovec, J.}
\newblock \bibinfo{title}{{BioSNAP Datasets}: {Stanford} biomedical network
  dataset collection}.
\newblock \bibinfo{howpublished}{\url{http://snap.stanford.edu/biodata}}
  (\bibinfo{year}{2018}).

\bibitem{tatonetti2012data}
\bibinfo{author}{Tatonetti, N.~P.}, \bibinfo{author}{Patrick, P.~Y.},
  \bibinfo{author}{Daneshjou, R.} \& \bibinfo{author}{Altman, R.~B.}
\newblock \bibinfo{title}{Data-driven prediction of drug effects and
  interactions}.
\newblock \emph{\bibinfo{journal}{Science translational medicine}}
  \textbf{\bibinfo{volume}{4}}, \bibinfo{pages}{125ra31--125ra31}
  (\bibinfo{year}{2012}).

\bibitem{zheng2019ddi}
\bibinfo{author}{Zheng, Y.} \emph{et~al.}
\newblock \bibinfo{title}{Ddi-pulearn: a positive-unlabeled learning method for
  large-scale prediction of drug-drug interactions}.
\newblock \emph{\bibinfo{journal}{BMC bioinformatics}}
  \textbf{\bibinfo{volume}{20}}, \bibinfo{pages}{1--12} (\bibinfo{year}{2019}).

\bibitem{wu2018moleculenet}
\bibinfo{author}{Wu, Z.} \emph{et~al.}
\newblock \bibinfo{title}{Moleculenet: a benchmark for molecular machine
  learning}.
\newblock \emph{\bibinfo{journal}{Chemical science}}
  \textbf{\bibinfo{volume}{9}}, \bibinfo{pages}{513--530}
  (\bibinfo{year}{2018}).

\bibitem{mousavian2014drug}
\bibinfo{author}{Mousavian, Z.} \& \bibinfo{author}{Masoudi-Nejad, A.}
\newblock \bibinfo{title}{Drug--target interaction prediction via chemogenomic
  space: learning-based methods}.
\newblock \emph{\bibinfo{journal}{Expert opinion on drug metabolism \&
  toxicology}} \textbf{\bibinfo{volume}{10}}, \bibinfo{pages}{1273--1287}
  (\bibinfo{year}{2014}).

\bibitem{chen2018machine}
\bibinfo{author}{Chen, R.}, \bibinfo{author}{Liu, X.}, \bibinfo{author}{Jin,
  S.}, \bibinfo{author}{Lin, J.} \& \bibinfo{author}{Liu, J.}
\newblock \bibinfo{title}{Machine learning for drug-target interaction
  prediction}.
\newblock \emph{\bibinfo{journal}{Molecules}} \textbf{\bibinfo{volume}{23}},
  \bibinfo{pages}{2208} (\bibinfo{year}{2018}).

\bibitem{wen2017deep}
\bibinfo{author}{Wen, M.} \emph{et~al.}
\newblock \bibinfo{title}{Deep-learning-based drug--target interaction
  prediction}.
\newblock \emph{\bibinfo{journal}{Journal of proteome research}}
  \textbf{\bibinfo{volume}{16}}, \bibinfo{pages}{1401--1409}
  (\bibinfo{year}{2017}).

\bibitem{Tsubaki2019}
\bibinfo{author}{Tsubaki, M.}, \bibinfo{author}{Tomii, K.} \&
  \bibinfo{author}{Sese, J.}
\newblock \bibinfo{title}{{Compound-protein interaction prediction with
  end-to-end learning of neural networks for graphs and sequences}}.
\newblock \emph{\bibinfo{journal}{Bioinformatics}}
  \textbf{\bibinfo{volume}{35}}, \bibinfo{pages}{309--318}
  (\bibinfo{year}{2019}).

\bibitem{heinonen2012metabolite}
\bibinfo{author}{Heinonen, M.}, \bibinfo{author}{Shen, H.},
  \bibinfo{author}{Zamboni, N.} \& \bibinfo{author}{Rousu, J.}
\newblock \bibinfo{title}{Metabolite identification and molecular fingerprint
  prediction through machine learning}.
\newblock \emph{\bibinfo{journal}{Bioinformatics}}
  \textbf{\bibinfo{volume}{28}}, \bibinfo{pages}{2333--2341}
  (\bibinfo{year}{2012}).

\bibitem{segler17}
\bibinfo{author}{Segler, M. H.~S.} \& \bibinfo{author}{Waller, M.~P.}
\newblock \bibinfo{title}{Modelling chemical reasoning to predict and invent
  reactions}.
\newblock \emph{\bibinfo{journal}{Chemistry: A European Journal}}
  \textbf{\bibinfo{volume}{23}}, \bibinfo{pages}{6118--6128}
  (\bibinfo{year}{2017}).

\bibitem{vaswani2017attention}
\bibinfo{author}{Vaswani, A.} \emph{et~al.}
\newblock \bibinfo{title}{Attention is all you need}.
\newblock In \emph{\bibinfo{booktitle}{Advances in neural information
  processing systems}}, \bibinfo{pages}{5998--6008} (\bibinfo{year}{2017}).

\bibitem{hendrycks2016gaussian}
\bibinfo{author}{Hendrycks, D.} \& \bibinfo{author}{Gimpel, K.}
\newblock \bibinfo{title}{Gaussian error linear units (gelus)}.
\newblock \emph{\bibinfo{journal}{arXiv preprint arXiv:1606.08415}}
  (\bibinfo{year}{2016}).

\bibitem{mnih2013learning}
\bibinfo{author}{Mnih, A.} \& \bibinfo{author}{Kavukcuoglu, K.}
\newblock \bibinfo{title}{Learning word embeddings efficiently with
  noise-contrastive estimation}.
\newblock In \emph{\bibinfo{booktitle}{Advances in neural information
  processing systems}}, \bibinfo{pages}{2265--2273} (\bibinfo{year}{2013}).

\end{thebibliography}
